\documentclass[10pt,twocolumn,letterpaper]{article}

\usepackage{wacv}
\usepackage{times}
\usepackage{epsfig}
\usepackage{graphicx}
\usepackage{amsmath}
\usepackage{amssymb}

\usepackage{subcaption}
\usepackage{algorithm,algpseudocode}
\usepackage{bbm}
\usepackage{adjustbox}
\usepackage[font=small]{caption}

\def\wacvPaperID{1276} % Enter the WACV Paper ID here

\wacvfinalcopy % *** Uncomment this line for the final submission

\ifwacvfinal
\def\assignedStartPage{9876} % *** Enter the assigned starting page number (instead of 9876)
\fi

\ifwacvfinal
\usepackage[breaklinks=true,bookmarks=false]{hyperref}
\else
\usepackage[pagebackref=true,breaklinks=true,colorlinks,bookmarks=false]{hyperref}
\fi

\ifwacvfinal
\else
\fi

\begin{document}
\pagenumbering{gobble}
%%%%%%%%% TITLE
\title{ClassMix: Segmentation-Based Data Augmentation for Semi-Supervised Learning}

\author{Viktor Olsson$^{1,2}$\thanks{Equal contribution.},
Wilhelm Tranheden$^{1,2}${\footnotemark[\value{footnote}]},
Juliano Pinto$^{1}$,
Lennart Svensson$^{1}$\\
$^1$Chalmers University of Technology, Gothenburg, Sweden\\
$^2$Volvo Cars, Gothenburg, Sweden\\
{\tt\small \{viktor.olsson.3,wilhelm.tranheden\}@volvocars.com, \{juliano,lennart.svensson\}@chalmers.se}
%\and
}

\maketitle
%\thispagestyle{empty}

%%%%%%%%% ABSTRACT
\begin{abstract}
   The state of the art in semantic segmentation is steadily increasing in performance, resulting in more precise and reliable segmentations in many different applications. However, progress is limited by the cost of generating labels for training, which sometimes requires hours of manual labor for a single image. Because of this, semi-supervised methods have been applied to this task, with varying degrees of success. A key challenge is that common augmentations used in semi-supervised classification are less effective for semantic segmentation. We propose a novel data augmentation mechanism called ClassMix, which generates augmentations by mixing unlabelled samples, by leveraging on the network's predictions for respecting object boundaries. We evaluate this augmentation technique on two common semi-supervised semantic segmentation benchmarks, showing that it attains state-of-the-art results. Lastly, we also provide extensive ablation studies comparing different design decisions and training regimes. 
\end{abstract}

%\footnotetext[*]{Done as part of a Master thesis at Volvo Cars}

%%%%%%%%% BODY TEXT
\section{Introduction}
Semantic segmentation is the task of assigning a semantic label to each pixel of an image. This is an essential part of many applications such as autonomous driving, medical imaging and scene understanding. Significant progress has been made in the area based on fully convolutional network architectures \cite{long2014fully,DeepLabv1,PSPNet}. When training deep learning models for semantic segmentation, a common bottleneck is the availability of ground-truth labels. In contrast, unlabelled data is usually abundant, and effectively leveraging it has the potential to increase performance with low cost. 

Semi-supervised learning based on consistency regularization has recently seen remarkable progress for image classification \cite{xie2019unsupervised,FixMatch}, utilizing strong data augmentations to enforce consistent predictions on unlabelled images. Augmentation techniques commonly used in classification have however proved ineffective for semi-supervised semantic segmentation \cite{French,ouali2020semisupervised}. Recent works have addressed this issue by either applying perturbations on an encoded state of the network instead of the input \cite{ouali2020semisupervised}, or by using the augmentation technique CutMix \cite{CutMix} to enforce consistent predictions over mixed samples \cite{French,StructuredLoss}. 

We propose a segmentation-based data augmentation strategy, ClassMix, and describe how it can be used for semi-supervised semantic segmentation. The augmentation strategy cuts half of the predicted classes from one image and pastes them onto another image, forming a new sample that better respects semantic boundaries while remaining accessible without ground-truth annotations. This is achieved by exploiting the fact that the network learns to predict a pixel-level semantic map of the original images. The predictions on mixed images are subsequently trained to be consistent with predictions made on the images before mixing. Following a recent trend in state-of-the-art consistency regularization for classification we also integrate entropy minimization \cite{xie2019unsupervised,FixMatch,berthelot2019mixmatch,ReMixMatch}, encouraging the network to generate predictions with low entropy on unlabelled data. We use pseudo-labelling \cite{pseudo-label} to accomplish this, and provide further motivations for combining it with ClassMix. Our proposed method is evaluated on established benchmarks for semi-supervised semantic segmentation, and an ablation study is included, analyzing the individual impact of different design and experimental choices.

Our main contributions can be summarised as: (1) We introduce an augmentation strategy which is novel for semantic segmentation, which we call ClassMix. (2) We incorporate ClassMix in a unified framework that makes use of consistency regularization and pseudo-labelling for semantic segmentation. (3) We demonstrate the effectiveness of our method by achieving state-of-the-art results in semi-supervised learning for the Cityscapes dataset \cite{Cityscapes}, as well as competitive results for the Pascal VOC dataset \cite{pascal-voc-2012}. Code is available at \url{https://github.com/WilhelmT/ClassMix}.

\section{Related Work}

%\subsubsection{Semantic segmentation}

%\subsubsection{Semi-supervised learning.}
For semantic segmentation, semi-supervised learning has been explored with techniques based on adversarial learning \cite{Hung,Mittal,qi2019ke}, consistency regularization \cite{French,StructuredLoss,medicalseg1,WeightAveragedConsistency}, and pseudo-labelling \cite{Feng,chen2020leveraging}. Our proposed method primarily incorporates ideas from the latter two approaches, which are expanded upon in subsequent sections.

%Various approaches also try to exploit the use of weak annotations, in a setting called weakly supervised learning \cite{ouali2020semisupervised,wang2018weaklysupervised,ibrahim2018semisupervised}. We, however, operate strictly in the semi-supervised learning setting and do not describe these methods further.

%\subsection{Adversarial learning}
%Until recently, semi-supervised semantic segmentation was dominated by GAN-based adversarial learning. Hung et al. \cite{Hung}, as well as Mittal et al. \cite{Mittal}, view the segmentation network as a generator network and train adversarially against a discriminator network, encouraging realistic semantic maps to be predicted for unlabelled images.

\subsection{Consistency Regularization}
The core idea in consistency regularization is that predictions for unlabelled data should be invariant to perturbations. A popular technique for classification is augmentation anchoring \cite{xie2019unsupervised,FixMatch,ReMixMatch}, where predictions performed on strongly augmented samples are enforced to follow predictions on weakly augmented versions of the same images. Our method utilizes augmentation anchoring in that consistency is enforced from unperturbed images to mixed images. Mixing images will create occlusions and classes in difficult contexts, hence being a strong augmentation.

Until recently, consistency regularization had been successfully applied for semantic segmentation only in the context of medical imaging \cite{medicalseg1,WeightAveragedConsistency}. Researchers pointed out the difficulties of performing consistency regularization for semantic segmentation, such as the violation of the cluster assumption, as described in \cite{French,ouali2020semisupervised}. Ouali et al. \cite{ouali2020semisupervised} propose to apply perturbations to the encoder's output, where the cluster assumption is shown to hold. Other approaches \cite{French,StructuredLoss} instead use a data augmentation technique called CutMix \cite{CutMix}, which composites new images by mixing two original images, resulting in images with some pixels from one image and some pixels from another image. Our proposed method, ClassMix, builds upon this line of research by using predictions of a segmentation network to construct the mixing. In this way, we can enforce consistency over highly varied mixed samples while a the same time better respecting the semantic boundaries of the original images.

\subsection{Pseudo-labelling}
Another technique used for semi-supervised learning is pseudo-labelling, training against targets based on the network class predictions, first introduced in \cite{pseudo-label}. Its primary motivation comes from entropy minimization, to encourage the network to perform confident predictions on unlabelled images. Such techniques have shown recent success in semi-supervised semantic segmentation \cite{Feng,chen2020leveraging,kalluri2019universal}. Some methods of semi-supervised learning for classification \cite{xie2019unsupervised,FixMatch,berthelot2019mixmatch,ReMixMatch} integrate entropy minimization in the consistency regularization framework. This is achieved by having consistency targets either sharpened \cite{xie2019unsupervised,berthelot2019mixmatch,ReMixMatch} or pseudo-labelled \cite{FixMatch}. Our proposed method of consistency regularization also naturally incorporates pseudo-labelling, as it prevents predictions close to mixing borders being trained to unreasonable classes, which will be further explained in coming sections.

\subsection{Related Augmentation Strategies}
In the CutMix algorithm \cite{CutMix}, randomized rectangular regions are cut out from one image and pasted onto another. This technique is based on mask-based mixing, where two images are mixed using a binary mask of the same size as the images. Our proposed technique, ClassMix, is based on a similar principle of combining images and makes use of predicted segmentations to generate the binary masks, instead of rectangles.

ClassMix also shares similarities with other segmentation-based augmentation strategies \cite{dwibedi2017cut,remez2018learning,Dvornik2018,Tripathi_2019_CVPR,Fang_2019_ICCV}, where annotated single instances of objects are cut out of images, and pasted onto new background scenes. Our way of combining two images conditioned on the predicted semantic maps exploits the same idea of compositing images. However, in contrast to several existing techniques our proposed augmentation does not rely on access to ground-truth segmentation-masks, allowing us to learn from unlabelled images in the semi-supervised setting. Additionally, we perform semantic segmentation with multiple classes present in each image rather than single instances, allowing variety by randomizing which classes to transfer. As previously mentioned, ClassMix is formulated as a generalisation of CutMix, using a binary mask to mix two randomly sampled images. This means that we only distinguish between foreground and background when generating the binary mask and not when training on the mixed images. Segmenting both foreground and background images means that semantic objects recognized by the network do not only have to be invariant to their context, but invariant to a diverse set of occlusions as well. %Lastly, our goal with using ClassMix is to enforce consistency over mixed and non-mixed unlabelled images for semi-supervised learning.

%===========================================================
\section{Method}
This section describes the proposed approach for semi-supervised semantic segmentation. It starts by explaining the data augmentation mechanism ClassMix, followed by a description of the loss function used, along with other details about the training procedure.

\subsection{ClassMix: Main Idea}
The proposed method performs semi-supervised semantic segmentation by using a novel data augmentation technique, ClassMix, which uses the unlabelled samples in the dataset to synthesize new images and corresponding artificial labels (``artificial label'' in this context is used to refer to the target that is used to train on the augmented image in the augmentation anchoring setup). ClassMix uses two unlabelled images as input and outputs a new augmented image, together with the corresponding artificial label for it. This augmented output is comprised of a mix of the inputs, where half of the semantic classes of one of the images are pasted on top of the other, resulting in an output which is novel and diverse, but still rather similar to the other images in the dataset.

\begin{figure*}[tp!]
    \centering
    \includegraphics[width=0.7\textwidth]{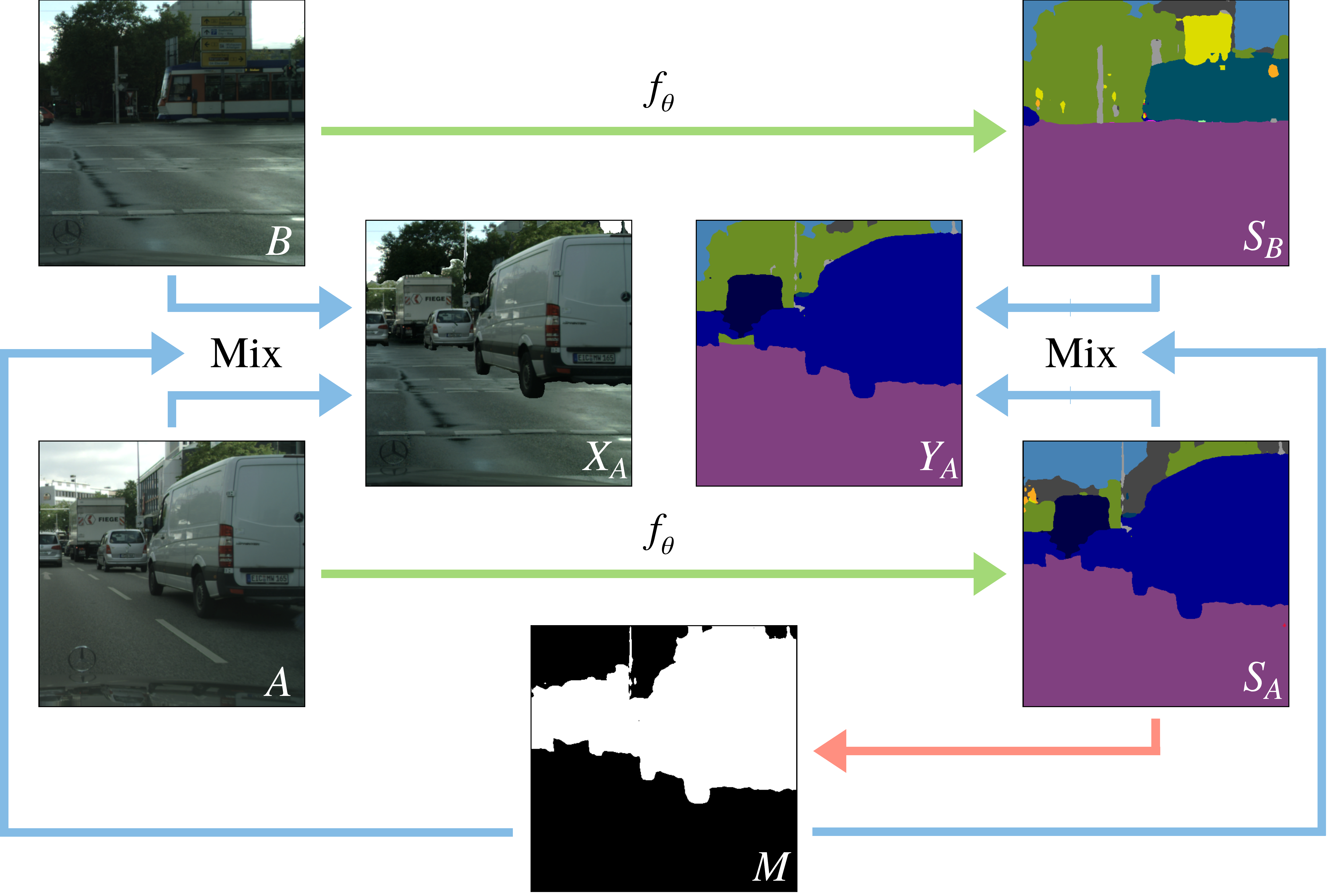}
    \caption{ClassMix augmentation technique. Two images $A$ and $B$ are sampled from the unlabelled dataset. Based on the prediction $S_A$ of image A, a binary mask $M$ is created. The mask is then used to mix the images $A$ and $B$ and their respective predictions into an augmented image $X_A$ and the corresponding artificial label $Y_A$.}
    \label{fig:classmix_diagram}
\end{figure*}

Figure \ref{fig:classmix_diagram} illustrates the essence of how ClassMix works. Two unlabelled images, $A$ and $B$, are sampled from the dataset. Both are fed through the segmentation network, $f_\theta$, which outputs the predictions $S_A$ and $S_B$. A binary mask $M$ is generated by randomly selecting half of the classes present in the argmaxed prediction $S_A$ and setting the pixels from those classes to have value 1 in $M$, whereas all others will have value 0. This mask is then used to mix images $A$ and $B$ into the augmented image $X_A$, which will contain pixels from $A$ where the mask had 1's and pixels from $B$ elsewhere. The same mixing is also done to the predictions $S_A$ and $S_B$, resulting in the artificial label $Y_A$. While artifacts may appear because of  the  nature  of  the  mixing  strategy, as  training  progresses  they  become fewer and smaller. Additionally, consistency regulation tends to yield good performance also with imperfect labels and this is further confirmed by our strong results.

\subsection{ClassMix: Details}
Two other techniques were added on top of the version of ClassMix presented in the previous section for improving its performance. This subsection explains those changes and provides a detailed description of the final ClassMix algorithm in pseudocode, in Algorithm \ref{alg:classmix}.

\begin{algorithm}[h]
\caption{ClassMix algorithm}
\label{alg:classmix}
\begin{algorithmic}[1]
\Require Two unlabelled samples $A$ and $B$, segmentation network $f_{\theta'}$.

%\State $S_A\gets \text{Pseudo-label}\Big(f_{\theta'}(A)\Big)$
%\State $S_B\gets \text{Pseudo-label}\Big(f_{\theta'}(B)\Big)$
\State $S_A\gets f_{\theta'}(A)$
\State $S_B\gets f_{\theta'}(B)$
\State $\tilde{S}_A \gets \arg\max_{c'}S_A(i,j,c')$ \Comment{Take pixel-wise argmax over classes.}
%\State $C\gets$ Set of the different classes present in $S_A$
\State $C\gets$ Set of the different classes present in $\tilde{S}_A$
\State $c\gets$ Randomly selected subset of $C$ such that $|c|=|C|/2$

\State For all $i,j$: $M(i,j)=\left\{\begin{matrix}
1 \text{, if } \tilde{S}_A(i,j)\in c
\\ 
\hspace{-16pt} 0 \text{, otherwise}
\end{matrix}\right.$ \Comment{Create binary mask.}

\State $X_A\gets M\odot A+(1-M)\odot B$ \Comment{Mix images.}
\State $Y_A\gets M\odot S_A+(1-M)\odot S_B$ \Comment{Mix predictions.}

\State \Return $X_A, Y_A$
\end{algorithmic}
\end{algorithm}

\noindent {\bf Mean-Teacher Framework.} In order to improve stability in the predictions for ClassMix, we follow a trend in state-of-the-art semi-supervised learning \cite{French,berthelot2019mixmatch,ict} and use the Mean Teacher Framework, introduced in \cite{TarvainenMeanTeacher}. Instead of using $f_\theta$ to make predictions for the inputs images $A$ and $B$ in ClassMix, we use $f_{\theta'}$, where $\theta'$ is an exponential moving average of the previous values of $\theta$ throughout the optimization. This type of temporal ensembling is cheap and simple to introduce in ClassMix, and results in more stable predictions throughout the training, and consequently more stable artificial labels for the augmented images. The network $f_{\theta}$ is then used to make predictions on the mixed images $X_A$, and the parameters $\theta$ are subsequently updated using gradient descent.% using backpropagation.

\vspace{0.4cm}
\noindent {\bf Pseudo-labelled Output.}
Another important detail about ClassMix is that, when generating labels for the augmented image, the artificial label $Y_A$ is ``argmaxed''. That is, the probability mass function over classes for each pixel is changed to a one-hot vector with a one in the class which was assigned the highest probability, zero elsewhere. This forms a pseudo-label to be used in training, and it is a commonly used technique in semi-supervised learning in order to encourage the network to perform confident predictions.

\begin{figure}[tp!]
    \centering
    \begin{subfigure}[b]{0.3\columnwidth}
        \includegraphics[width=\textwidth]{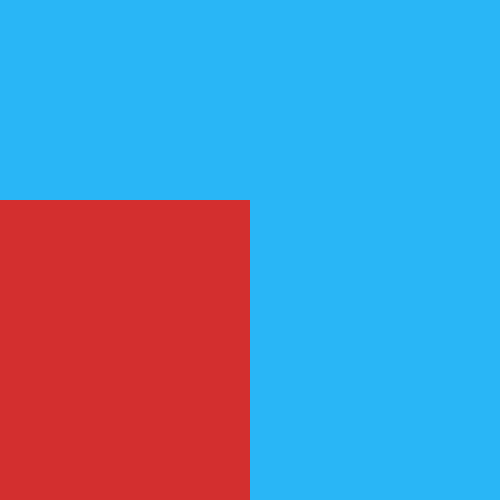}
    \end{subfigure}
    ~
    \begin{subfigure}[b]{0.3\columnwidth}
        \includegraphics[width=\textwidth]{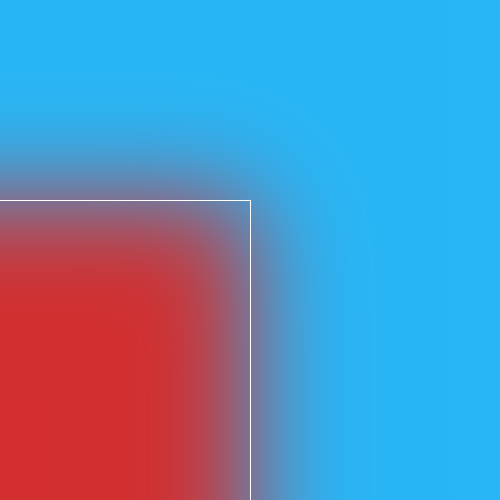}
    \end{subfigure}
    ~
    \begin{subfigure}[b]{0.3\columnwidth}
        \includegraphics[width=\textwidth]{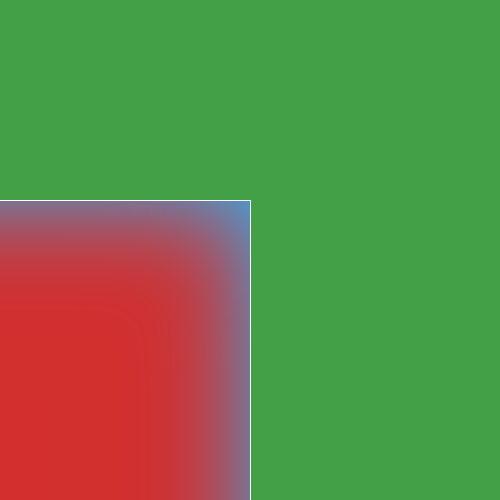}
    \end{subfigure}
    \caption{Toy example of label contamination with 3 different classes. Left: Ground-truth labels; red is class 1, blue is class 2. Middle: Prediction made by network; regions where the network is uncertain between classes 1 and 2 have a mix of red and blue colors. The decision boundary is marked with a white line. Right: The red class is pasted on top of a new image, which is comprised entirely of the third class. Note how the pasted class still brings some uncertainty of class 2 (blue) to the new image. This results in problematic artificial labels for training, since the context around the pasted object now changed to class 3.}
    \label{fig:label_contamination}
\end{figure}

For ClassMix, pseudo-labelling serves an additional purpose, namely eliminating uncertainty along borders. Since the mask $M$ is generated from the output prediction of $A$, the edges of the mask will be aligned with the decision boundaries of the semantic map. This comes with the issue that predictions are especially uncertain close to the mixing borders, as the segmentation task is hardest close to class boundaries \cite{li2017pixels}. This results in a problem we denote label contamination, illustrated in figure \ref{fig:label_contamination}. When the classes chosen by $M$ are pasted on top of image $B$, their adjacent context will often change, resulting in poor artificial labels. Pseudo-labelling effectively mitigates this issue, since the probability mass function for each pixel is changed to a one-hot vector for the most likely class, therefore ``sharpening'' the artificial labels, resulting in no contamination.
%We further note that even when predictions are not perfect, label contamination is still an important issue as its actually arising at the decision boundaries, which the mask is aligned to, not the actual true semantic boundaries.

 %Because the boundary of the mask generated from image $A$ aligns with the semantic boundaries of image $A$, where the predictions from the network are especially uncertain between the two adjacent classes \cite{}, the adjacent class effectively ``contaminates'' the predictions, as illustrated in Figure \ref{fig:label_contamination}. When ClassMix pastes this part of image $A$ in $B$, the surrounding context of the contaminated label will possibly change, yielding a poor artificial label for the augmented image. 

\subsection{Loss and Training}

For all the experiments in this paper, we train the parameters of the semantic segmentation network $f_\theta$ by minimizing the following loss:
\begin{equation}
    L(\theta) = \mathbb E\left[\ell\Big(f_\theta(X_L), Y_L\Big) + \lambda \ell\Big(f_\theta(X_A), Y_A\Big)\right]~.
\end{equation}
In this expectation, $X_L$ is an image sampled uniformly at random from the dataset of labelled images, and $Y_L$ is its corresponding ground-truth semantic map. The random variables $X_A$ and $Y_A$ are respectively the augmented image and its artificial label, produced by the ClassMix augmentation method (as described in algorithm \ref{alg:classmix}), where the input images $A$ and $B$ are sampled uniformly at random from the unlabelled dataset. (in practice the augmentations are computed by mixing all the images within a batch, for efficiency reasons; we refer interested readers to our code for further details). Lastly, $\lambda$ is a hyper-parameter that controls the balance between the supervised and unsupervised terms, and $\ell$ is the cross-entropy loss, averaged over all pixel positions in the semantic maps, i.e.
\begin{equation}
    \ell(S, Y) = -\frac{1}{W\cdot H}\sum_{i=1}^W \sum_{j=1}^H \left( \sum_{c=1}^C Y(i,j,c)\cdot \log S(i,j,c) \right),
\end{equation}
where $W$ and $H$ are the width and height of the images, and $S(i,j,c)$ and $Y(i,j,c)$ are the probabilities that the pixel in coordinates $i$, $j$ belongs to class $c$, according to the prediction $S$ and target $Y$, respectively. We train $\theta$ by stochastic gradient descent on this loss, imposing batches with 50\% labelled data and 50\% augmented data. 

It is beneficial to the training progress that the unsupervised weight $\lambda$ starts close to zero, because initially the network predictions are of low quality and therefore the pseudo-labels generated will not be reasonable targets for training on the augmented images. As the predictions of the network improve, this weight can then be increased. 
%This was accomplished by having an adaptive value of $\lambda$ that varies for each augmentation, according to a measure of how certain the network's predictions are for the corresponding unlabelled samples used to generate it. 
This was accomplished by setting the value of $\lambda$ for an augmented sample as the proportion of pixels in its artificial label where the probability of the most likely class is above a predetermined threshold $\tau$. This results in a value between 0 and 1, which we empirically found to serve as an adequate proxy for the quality of the predictions, roughly in line with \cite{French,FrenchSelfEnsembling}. 
%We set $\lambda$ for an augmentation $X_A$ as the proportion of pixels in $A$ and $B$ (the unlabelled images used to generate the augmentation, as shown in Figure \ref{fig:classmix_diagram}) for which the certainty of the most likely class according to the network is above a given threshold $\tau$:
%\begin{equation}
%    \lambda(X_A) = \frac{1}{2\cdot W\cdot H}\sum_{i=1}^W\sum_{j=1}^H \mathbbm{1}\left(\max_c f_\theta(A)(i,j,c)>\tau\right)+\mathbbm{1}\left(\max_c f_\theta(B)(i,j,c)>\tau\right)
%\end{equation}
%where $\mathbbm{1}$ is the indicator function, and $f_\theta(I)(i,j,c)$ is the probability according to $f_\theta$ that the pixel in the $i,j$ position in image $I$ belongs to class $c$.

%===========================================================
\section{Experiments}
\label{sec:experiments}

In order to evaluate the proposed method, we perform experiments on two common semi-supervised semantic segmentation datasets, and this section presents the results obtained. Additionally, an extensive ablation study for motivating our design decisions is also provided, where we further investigate the properties of ClassMix and its components.

\subsection{Implementation Details and Datasets}
Our method is implemented using the PyTorch framework and training was performed on two Tesla V100 GPUs. We adopt the DeepLab-v2 framework \cite{DeepLabv2} with a ResNet101 backbone \cite{ResNet} pretrained on ImageNet \cite{imagenet} and MSCOCO \cite{mscoco}, identical to the ones used in \cite{Hung, Feng, Mittal}. 
As optimizer, we use Stochastic Gradient Descent with Nesterov acceleration, and a base learning rate of $2.5\times10^{-4}$, decreased with polynomial decay with power 0.9 as used in \cite{DeepLabv2}. Momentum is set to 0.9 and weight decay to $5\times 10^{-4}$.

We present results for two semantic segmentation datasets, Cityscapes \cite{Cityscapes} and Pascal VOC 2012 \cite{pascal-voc-2012}. The Cityscapes urban scenery dataset contains 2,975 training images and 500 validations images. We resize images to $512\times1024$ with no random cropping, scaling or flipping, use batches with 2 labelled and 2 unlabelled samples and train for 40k iterations, all in line with \cite{Hung}.
For the Pascal VOC 2012 dataset we use the original images along with the extra annotated images from the Semantic Boundaries dataset \cite{SemanticBoundaries}, resulting in 10,582 training images and 1,449 validation images. Images are randomly scaled between 0.5 and 1.5 as well as randomly horizontally flipped and after that cropped to a size of $321\times321$ pixels, also in line with \cite{Hung}. We train for 40k iterations using batches with 10 labelled and 10 unlabelled samples.

\begin{figure}[t]
    \centering
    \begin{adjustbox}{width=\columnwidth}
    \begin{subfigure}[b]{1.0\textwidth}
        \includegraphics[width=\textwidth]{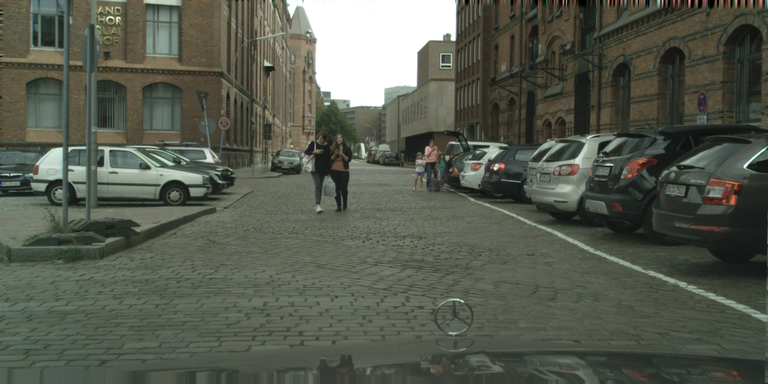}
    \end{subfigure}
    \hspace{0.01cm}
    \begin{subfigure}[b]{1.0\textwidth}
        \includegraphics[width=\textwidth]{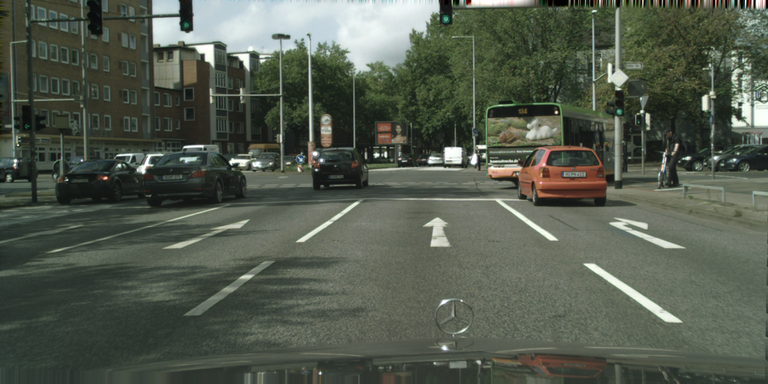}
    \end{subfigure}
    \hspace{0.01cm}
    \begin{subfigure}[b]{1.0\textwidth}
        \includegraphics[width=\textwidth]{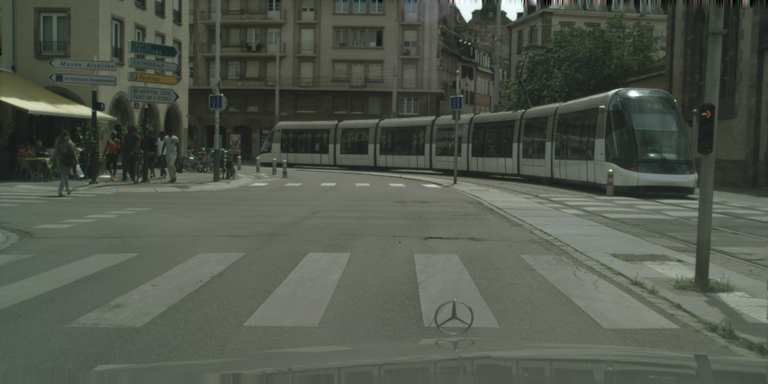}
    \end{subfigure}
    \end{adjustbox}
    
    \begin{adjustbox}{width=\columnwidth}
    \begin{subfigure}[b]{1.0\textwidth}
        \includegraphics[width=\textwidth]{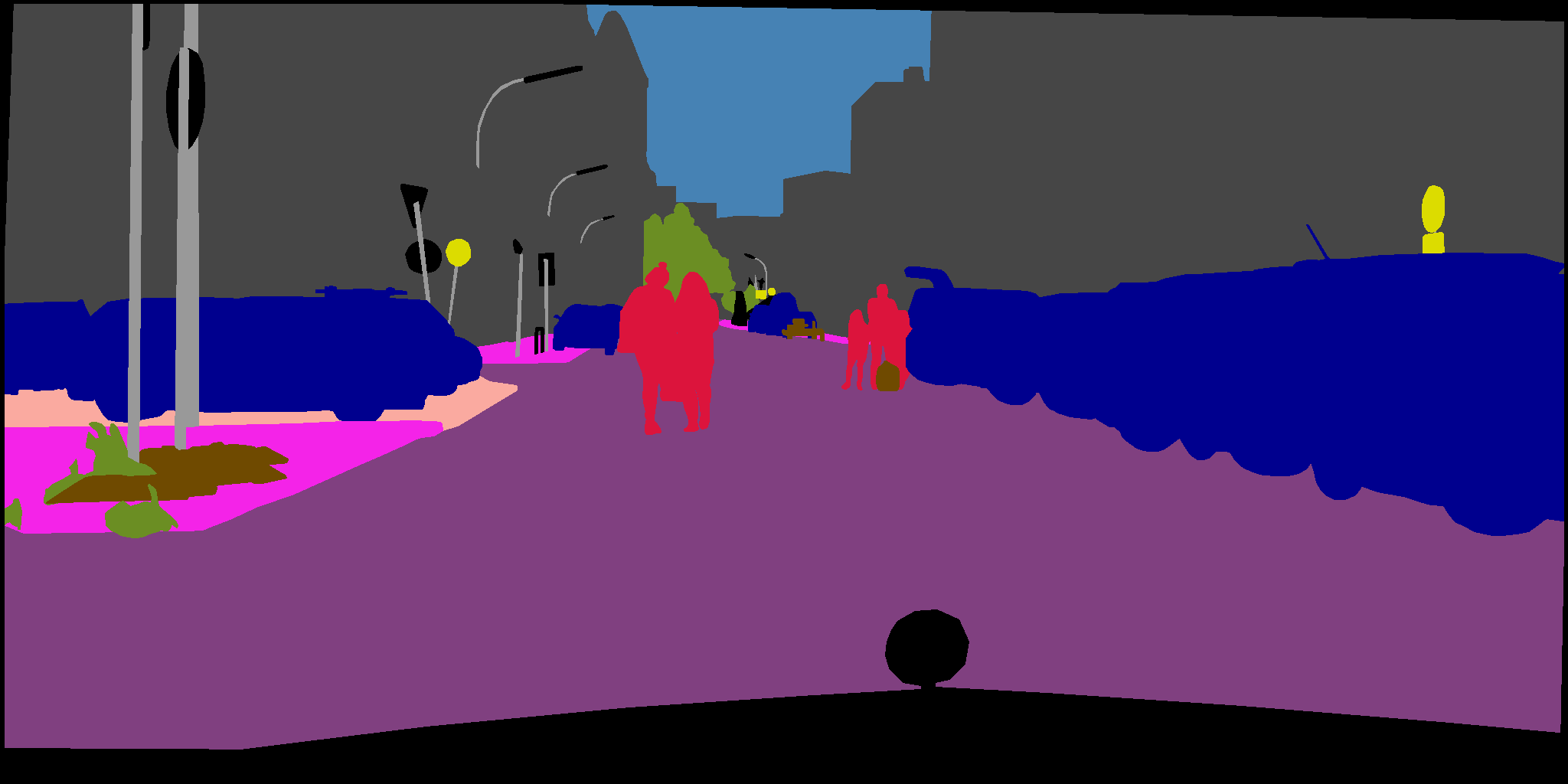}
    \end{subfigure}
    \hspace{0.01cm}
    \begin{subfigure}[b]{1.0\textwidth}
        \includegraphics[width=\textwidth]{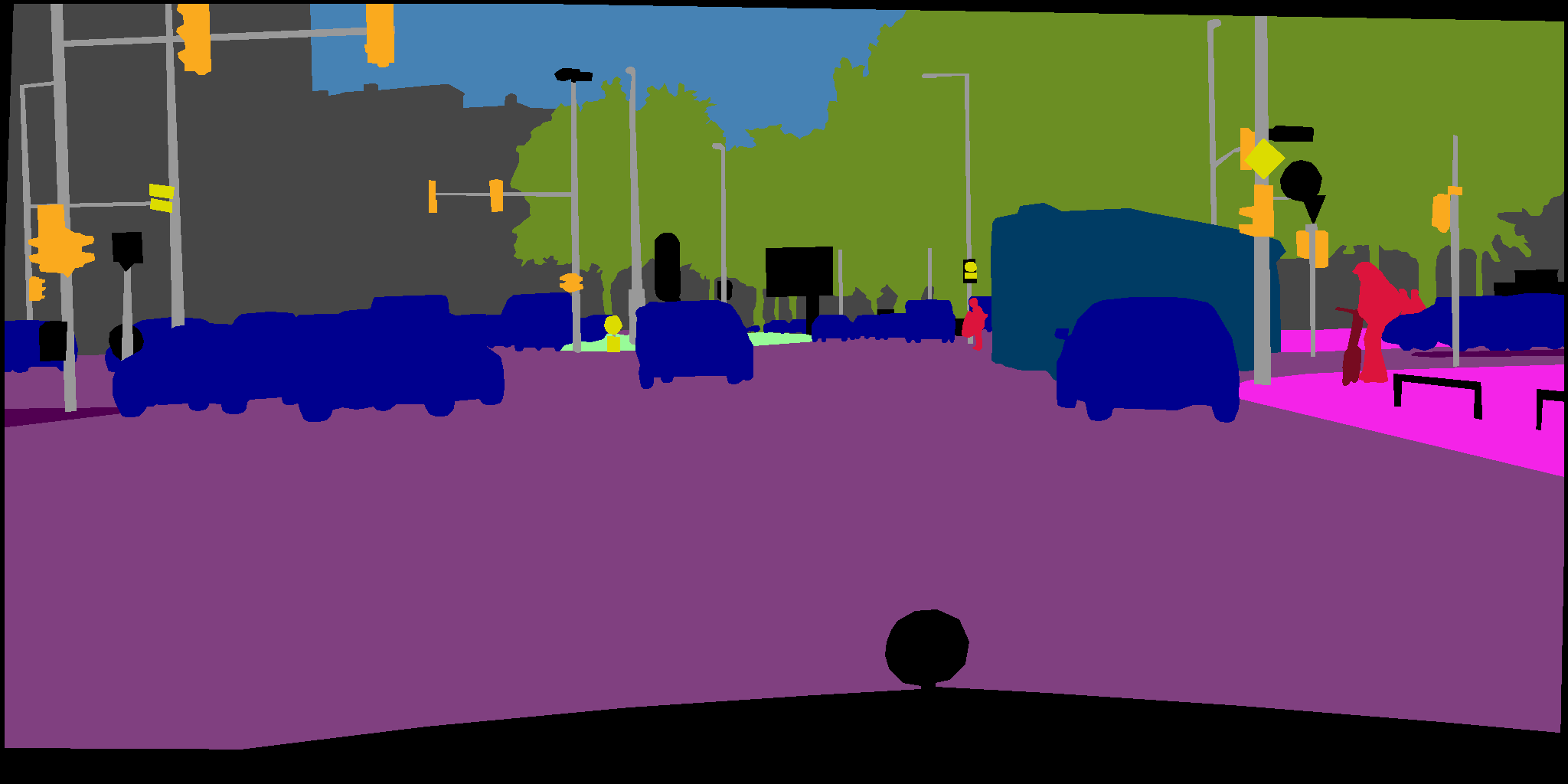}
    \end{subfigure}
    \hspace{0.01cm}
    \begin{subfigure}[b]{1.0\textwidth}
        \includegraphics[width=\textwidth]{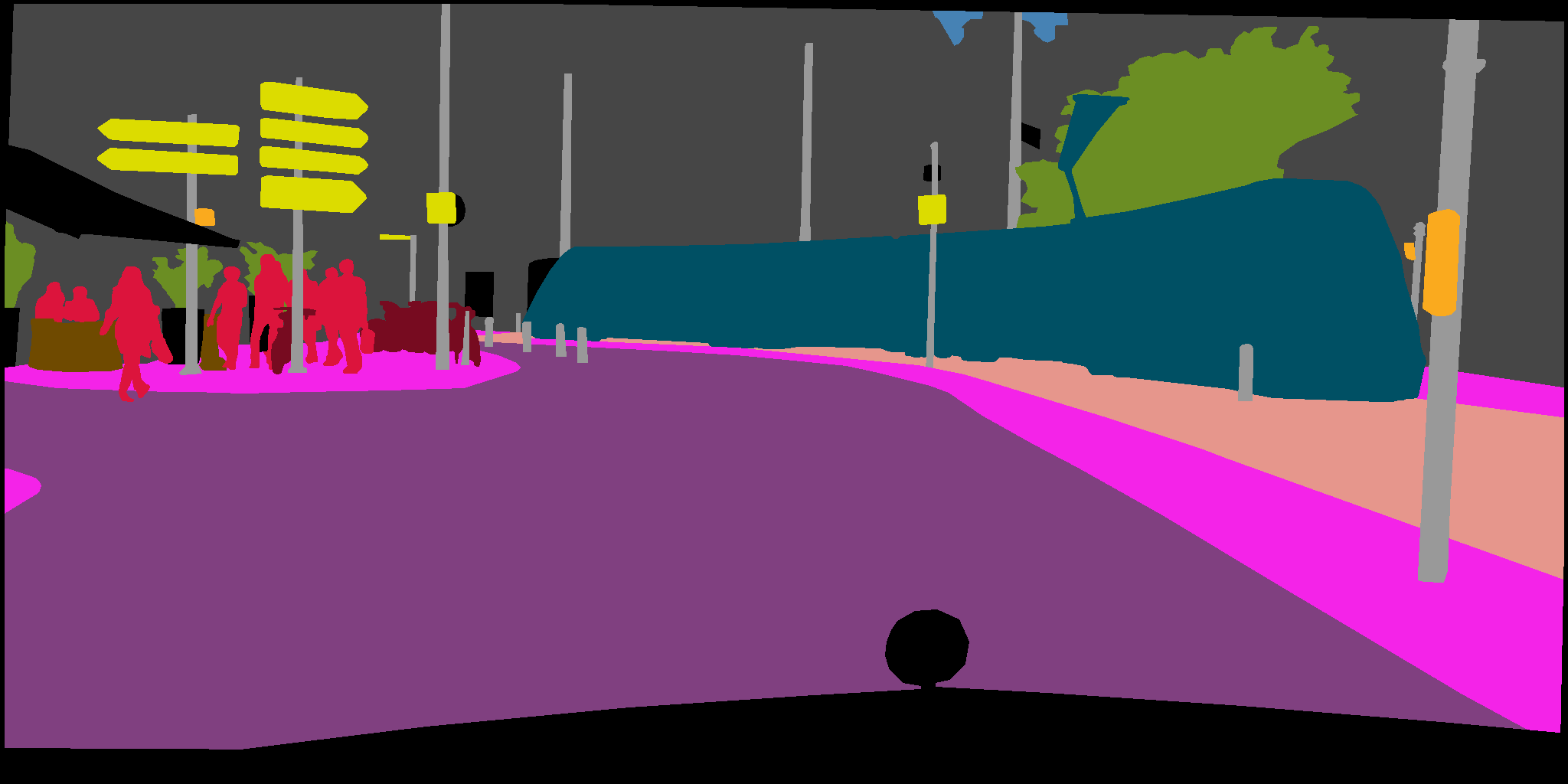}
    \end{subfigure}
    \end{adjustbox}
    \caption{Images and corresponding semantic maps from the Cityscapes dataset.}
    \label{fig:cityscapes}
\end{figure}

\begin{figure}[t]
    \centering
    \begin{adjustbox}{width=\columnwidth}
    \begin{subfigure}[b]{0.6\textwidth}
        \includegraphics[width=\textwidth]{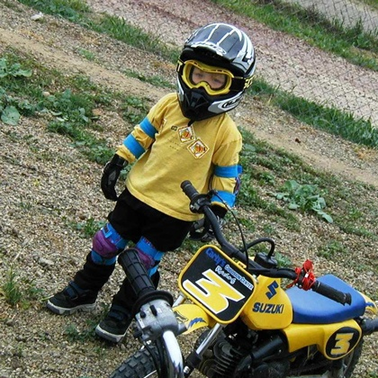}
    \end{subfigure}
    \hspace{0.01cm}
    \begin{subfigure}[b]{0.6\textwidth}
        \includegraphics[width=\textwidth]{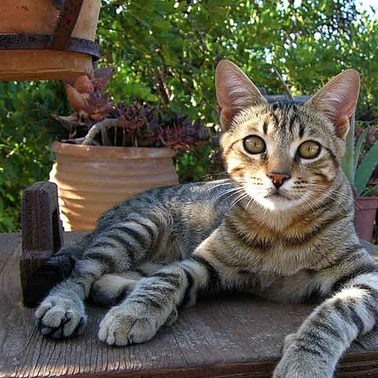}
    \end{subfigure}
    \hspace{0.01cm}
    \begin{subfigure}[b]{0.6\textwidth}
        \includegraphics[width=\textwidth]{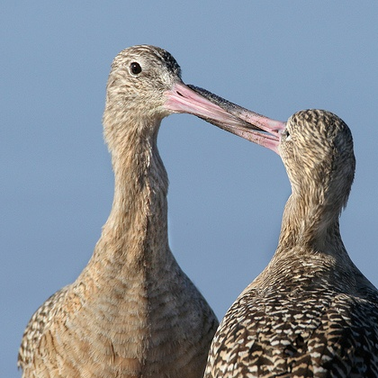}
    \end{subfigure}
    \hspace{0.01cm}
    \begin{subfigure}[b]{0.6\textwidth}
        \includegraphics[width=\textwidth]{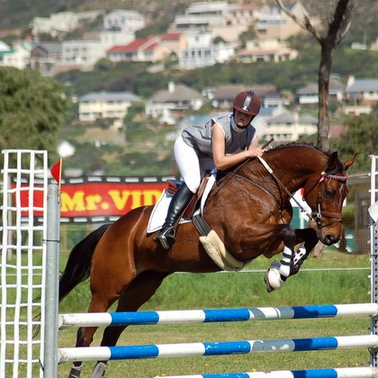}
    \end{subfigure}
    \hspace{0.01cm}
    \begin{subfigure}[b]{0.6\textwidth}
        \includegraphics[width=\textwidth]{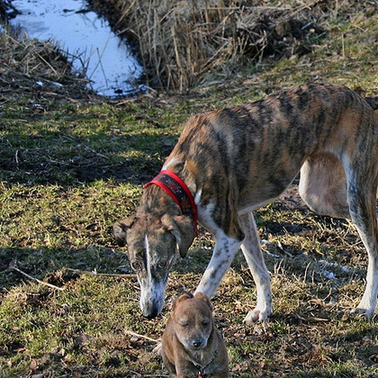}
    \end{subfigure}
    \end{adjustbox}
    \begin{adjustbox}{width=\columnwidth}
    \begin{subfigure}[b]{0.6\textwidth}
        \includegraphics[width=\textwidth]{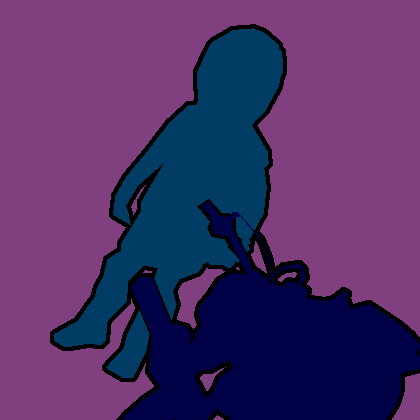}
    \end{subfigure}
    \hspace{0.01cm}
    \begin{subfigure}[b]{0.6\textwidth}
        \includegraphics[width=\textwidth]{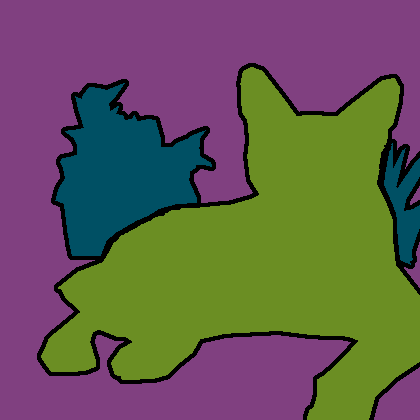}
    \end{subfigure}
    \hspace{0.01cm}
    \begin{subfigure}[b]{0.6\textwidth}
        \includegraphics[width=\textwidth]{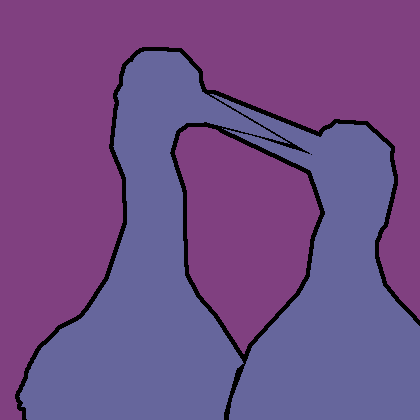}
    \end{subfigure}
    \hspace{0.01cm}
    \begin{subfigure}[b]{0.6\textwidth}
        \includegraphics[width=\textwidth]{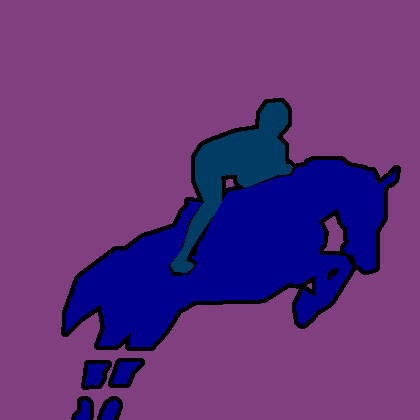}
    \end{subfigure}
    \hspace{0.01cm}
    \begin{subfigure}[b]{0.6\textwidth}
        \includegraphics[width=\textwidth]{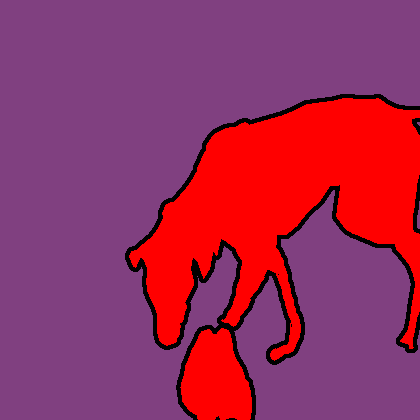}
    \end{subfigure}
    \end{adjustbox}
    
    \caption{Images and corresponding semantic maps from the Pascal VOC 2012 dataset.}
    \label{fig:pascal}
\end{figure}

Figures \ref{fig:cityscapes} and \ref{fig:pascal} show example images of both datasets along with their corresponding ground truth semantic maps. It is clear that the images in Cityscapes contain a lot more classes in each image than the Pascal images do. At the same time the semantic maps are more consistent throughout the images in the Cityscapes dataset than between Pascal images, for example the road and sky classes are almost always present and in approximately the same place.

%------------------------------------------------------------------------

\subsection{Results}
\label{subsec:results}

\noindent {\bf Cityscapes.} 
In Table \ref{tab:cityscapesresults} we present our results for the Cityscapes dataset, given as mean Intersection over Union (mIoU) scores. We have performed experiments for four proportions of labelled samples, which are given along with baselines that are trained in a supervised fashion on the corresponding data amounts. In the table we also provide results from four other papers, all using the same DeepLab-v2 framework. Hung et al. and Mittal et al. use an adversarial approach \cite{Hung, Mittal}, French et al. use consistency regularization \cite{French} and Feng et al. use a self-training scheme \cite{Feng}. We note that our results are higher for three out of four data amounts and that our improvement from the baseline result to the SSL result is higher for all amounts of training data.

\begin{table*}[t]
    \centering
    \caption{Performance (mIoU) on Cityscapes validation set averaged over three runs. Results from four previous papers are provided for comparison, all using the same DeepLab-v2 network with ResNet-101 backbone.}
    \begin{tabular}{ l l l l l l } 
        Labelled samples & 1/30  & 1/8  & 1/4  & 1/2  & Full (2975) \\
        \hline \hline
         
        Baseline & - & 55.5\% & 59.9\% & 64.1\% & 66.4\% \\
        Adversarial \cite{Hung} & - & 58.8\% & 62.3\% & 65.7\% & - \\
        Improvement & - & 3.3 & 2.4 & 1.6 & - \\
        \hline \hline
        
        Baseline & - & 56.2\% & 60.2\% & - & 66.0\% \\
        s4GAN \cite{Mittal}\footnotemark & - & 59.3\% & 61.9\% & - & 65.8\% \\
        Improvement & - & 3.1 & 1.7 & - & -0.2 \\
        \hline \hline
        
        Baseline & 44.41\% & 55.25\% & 60.57\% & - & 67.53\% \\
        French et al. \cite{French}\footnotemark[\value{footnote}] & 51.20\% & 60.34\% & 63.87\% & - & - \\
        Improvement & 6.79 & 5.09 & 3.3 & - & - \\
        \hline \hline
        
        Baseline & 45.5 \% & 56.7\% & 61.1\% & - & 66.9\% \\
        DST-CBC \cite{Feng} & 48.7 \% & 60.5\% & \textbf{64.4\%} & - & - \\
        Improvement & 3.2 & 3.8 & 3.3 & - & - \\
        \hline \hline
        
        Baseline & 43.84\%\scriptsize{$\pm$0.71} & 54.84\%\scriptsize{$\pm$1.14} & 60.08\%\scriptsize{$\pm$0.62} & 63.02\%\scriptsize{$\pm$0.14} & 66.19\%\scriptsize{$\pm$0.11} \\

        Ours & \textbf{54.07\%}\scriptsize{$\pm$1.61} & \textbf{61.35\%}\scriptsize{$\pm$0.62} & 63.63\%\scriptsize{$\pm$0.33} & \textbf{66.29\%}\scriptsize{$\pm$0.47} & - \\

        Improvement & \textbf{10.23} & \textbf{6.51} & \textbf{3.55} & \textbf{3.27} & - \\
        \hline
        
    \end{tabular}
    \label{tab:cityscapesresults}
\end{table*}{}

The fact that we achieve, to the best of our knowledge, the best SSL-results on the Cityscapes dataset further supports that consistency regularization can be successfully applied to semi-supervised semantic segmentation. French et al. use a method similar to ours \cite{French}, where they enforce consistency with CutMix as their mixing algorithm, instead of our ClassMix. We believe that one reason for the higher performance of ClassMix is the diversity of the masks created. This diversity stems from the fact that each image includes many classes and that each class often contains several objects. Since there are many classes in each image, an image rarely has the exact same classes being selected for mask generation several times, meaning that the masks based on a given image will be varied over the course of training. Furthermore, since each class often contains several objects, the masks will naturally become very irregular, and hence very different between images; when using CutMix, the masks will not be nearly as varied.

We believe that another reason that ClassMix works well is that the masks are based on the semantics of the images, as discussed previously. This minimizes the occurrence of partial objects in the mixed image, which are difficult to predict and make learning unnecessarily hard. It also means that mixing borders will be close to being aligned with boundaries of objects. This creates mixed images that better respect the semantic boundaries of the original images. They are consequently more realistic looking than images created using, e.g., CutMix, and lie closer to the underlying data distribution.

A third reason for ClassMix performing well for Cityscapes may be that images are similar within the dataset. All images have the road class in the bottom and sky at the top, and whenever there are cars or people, for example, they are roughly in the same place. Another way to put this is that classes are not uniformly distributed across the image area, but instead clustered in smaller regions of the image, as is shown by the spatial distribution of classes in Figure \ref{fig:spatialCS}.
%the spatial distribution of classes is not uniform, but instead rather low-entropic. 
Because of this, objects that are pasted from one image to another are likely to end up in a reasonable context, which may be important in the cases where objects are transferred without any surrounding context from the original image.

\footnotetext[1]{Same DeepLab-v2 network but with only ImageNet pre-training and not MSCOCO.}

\newcommand\imgsize{0.3}
\begin{figure}[h!]
    \centering
    
    \setlength{\fboxsep}{0pt}%
    \setlength{\fboxrule}{0.5pt}%
    \begin{subfigure}[b]{\imgsize\columnwidth}
        \captionsetup{justification=centering,skip=0pt}
        \fbox{\includegraphics[width=\textwidth]{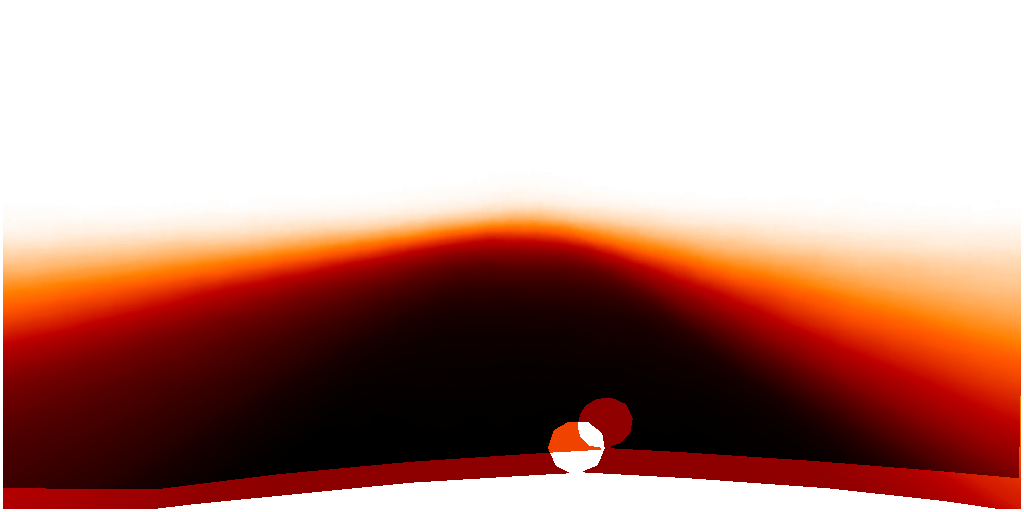}}
        \caption*{\footnotesize{Road}}
    \end{subfigure}
    \hspace{0.01cm}
    \begin{subfigure}[b]{\imgsize\columnwidth}
        \fbox{\includegraphics[width=\textwidth]{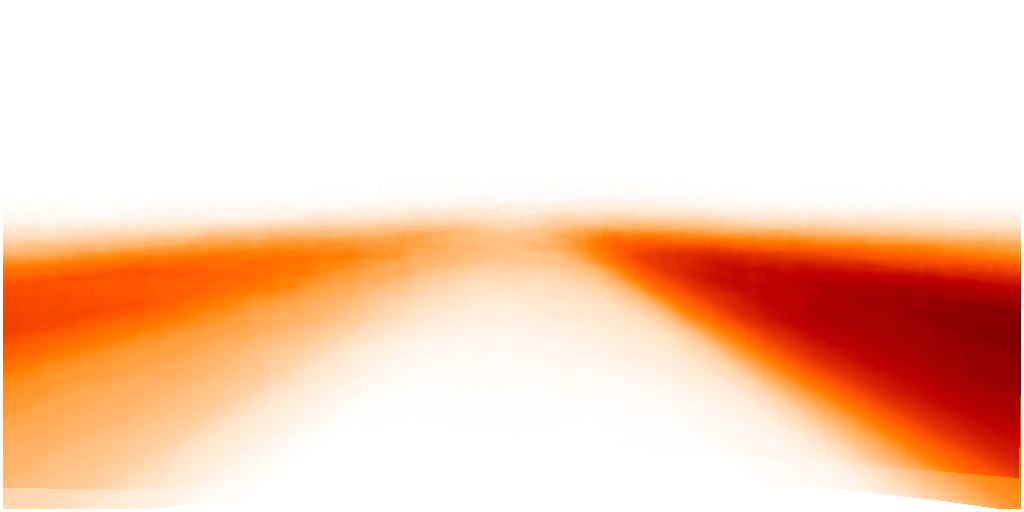}}
        \captionsetup{justification=centering,skip=0pt}
        \caption*{\footnotesize{Sidewalk}}
    \end{subfigure}
    \hspace{0.01cm}
    \begin{subfigure}[b]{\imgsize\columnwidth}
        \fbox{\includegraphics[width=\textwidth]{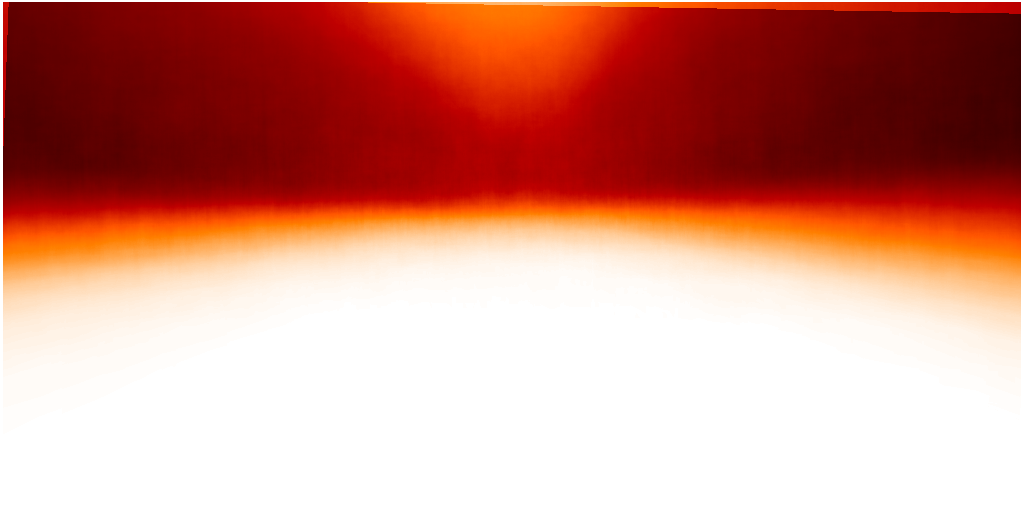}}
        \captionsetup{justification=centering,skip=0pt}
        \caption*{\footnotesize{Building}}
    \end{subfigure}
    \hspace{0.01cm}
    \begin{subfigure}[b]{\imgsize\columnwidth}
        \fbox{\includegraphics[width=\textwidth]{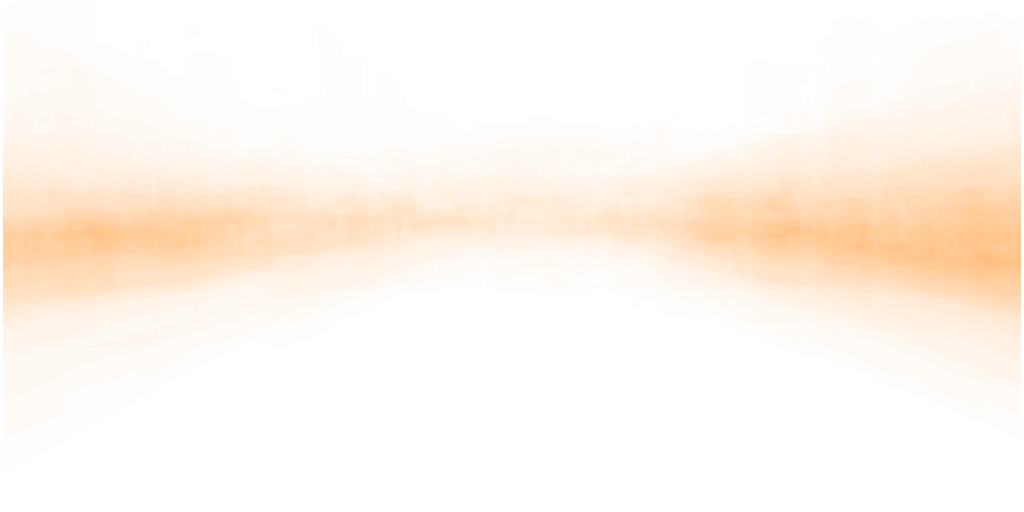}}
        \captionsetup{justification=centering,skip=0pt}
        \caption*{\footnotesize{Wall}}
    \end{subfigure}
    \hspace{0.01cm}
    \begin{subfigure}[b]{\imgsize\columnwidth}
        \fbox{\includegraphics[width=\textwidth]{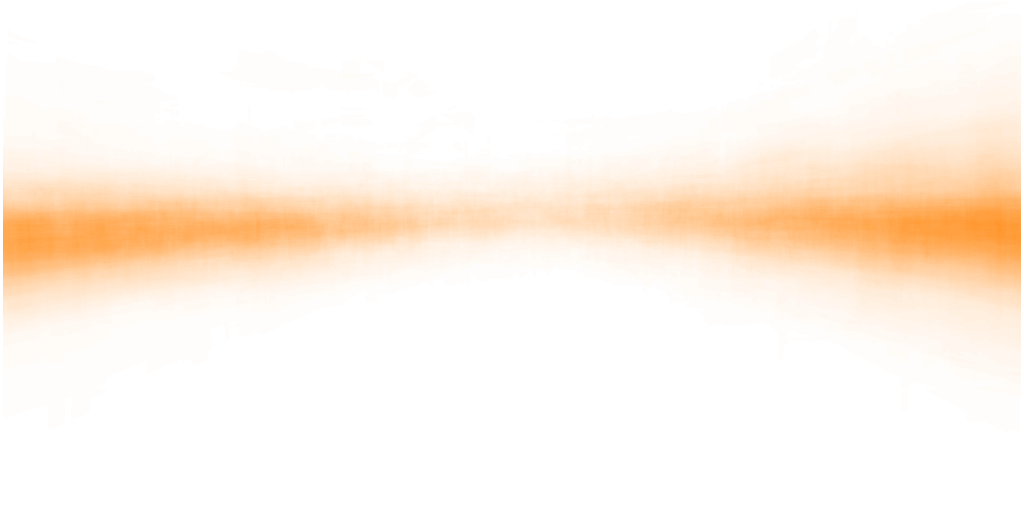}}
        \captionsetup{justification=centering,skip=0pt}
        \caption*{\footnotesize{Fence}}
    \end{subfigure}
    \hspace{0.01cm}
    \begin{subfigure}[b]{\imgsize\columnwidth}
        \fbox{\includegraphics[width=\textwidth]{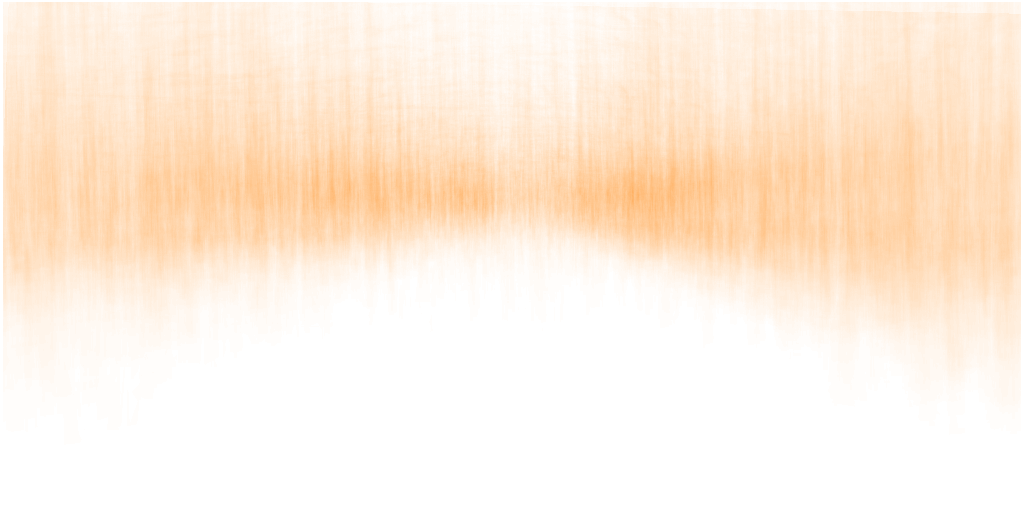}}
        \captionsetup{justification=centering,skip=0pt}
        \caption*{\footnotesize{Pole}}
    \end{subfigure}
    \hspace{0.01cm}
    \begin{subfigure}[b]{\imgsize\columnwidth}
        \fbox{\includegraphics[width=\textwidth]{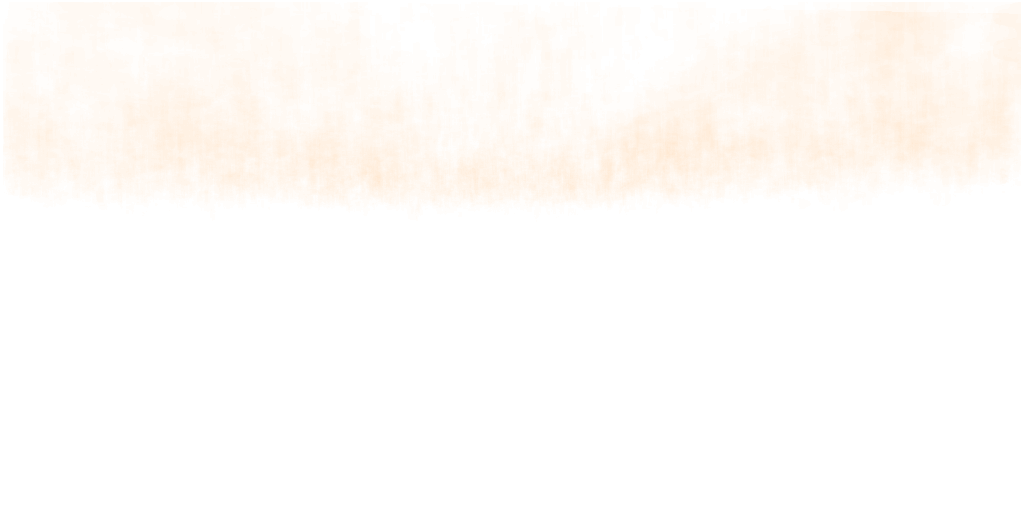}}
        \captionsetup{justification=centering,skip=0pt}
        \caption*{\footnotesize{Traffic light}}
    \end{subfigure}
    \hspace{0.01cm}
    \begin{subfigure}[b]{\imgsize\columnwidth}
        \fbox{\includegraphics[width=\textwidth]{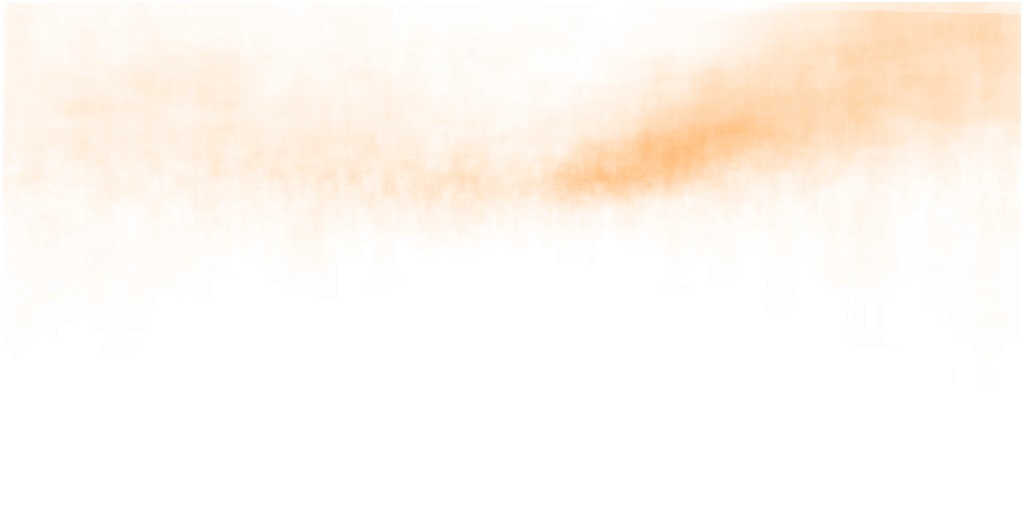}}
        \captionsetup{justification=centering,skip=0pt}
        \caption*{\footnotesize{Traffic sign}}
    \end{subfigure}
    \hspace{0.01cm}
    \begin{subfigure}[b]{\imgsize\columnwidth}
        \fbox{\includegraphics[width=\textwidth]{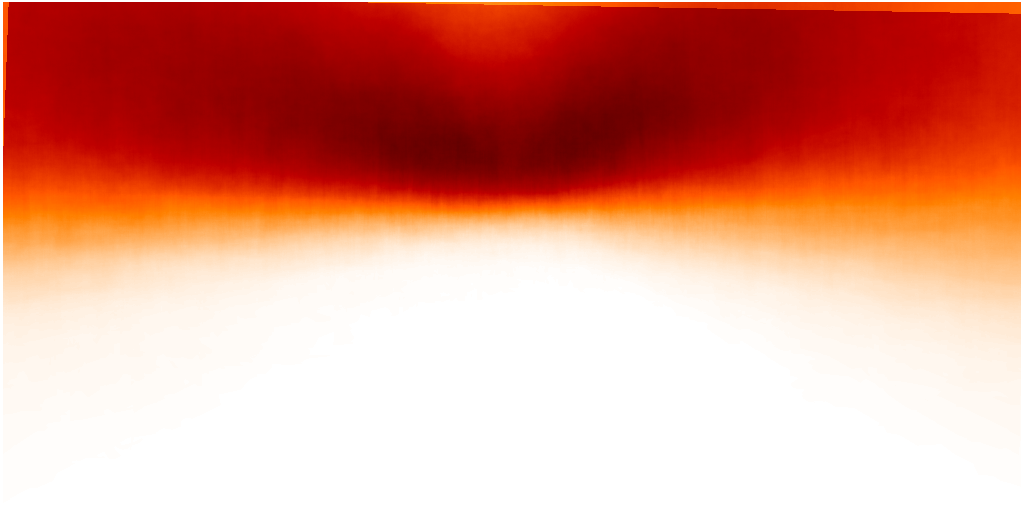}}
        \captionsetup{justification=centering,skip=0pt}
        \caption*{\footnotesize{Vegetation}}
    \end{subfigure}
    \hspace{0.01cm}
    \begin{subfigure}[b]{\imgsize\columnwidth}
        \fbox{\includegraphics[width=\textwidth]{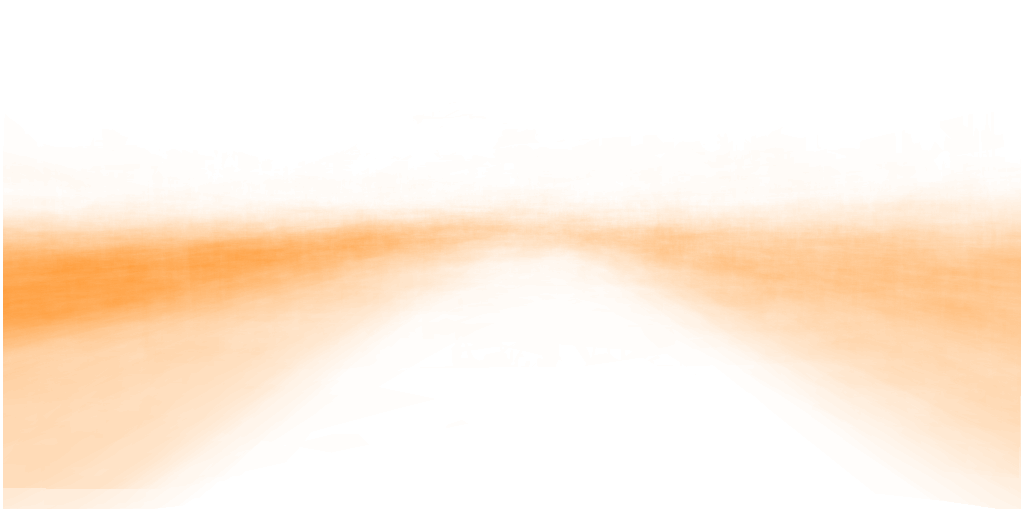}}
        \captionsetup{justification=centering,skip=0pt}
        \caption*{\footnotesize{Terrain}}
    \end{subfigure}
    \hspace{0.01cm}
    \begin{subfigure}[b]{\imgsize\columnwidth}
        \fbox{\includegraphics[width=\textwidth]{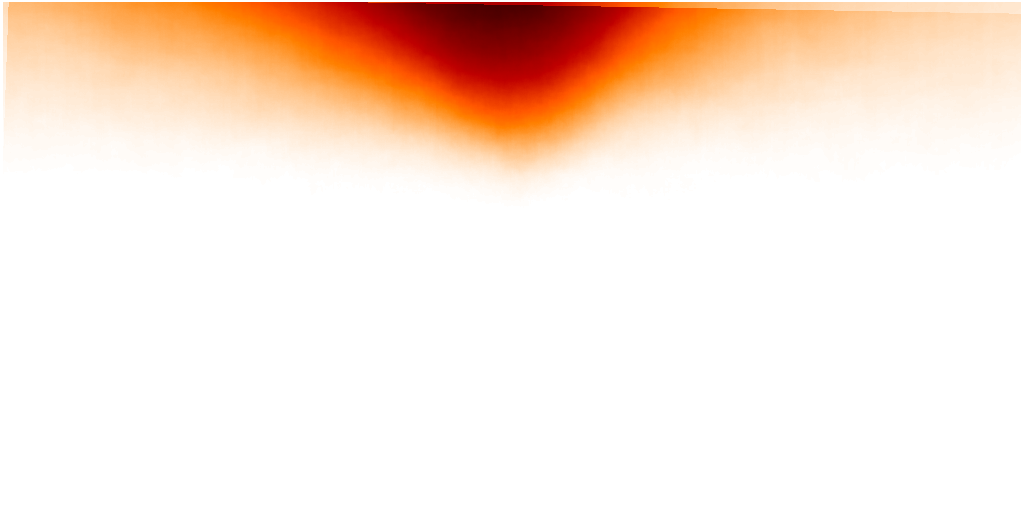}}
        \captionsetup{justification=centering,skip=0pt}
        \caption*{\footnotesize{Sky}}
    \end{subfigure}
    \hspace{0.01cm}
    \begin{subfigure}[b]{\imgsize\columnwidth}
        \fbox{\includegraphics[width=\textwidth]{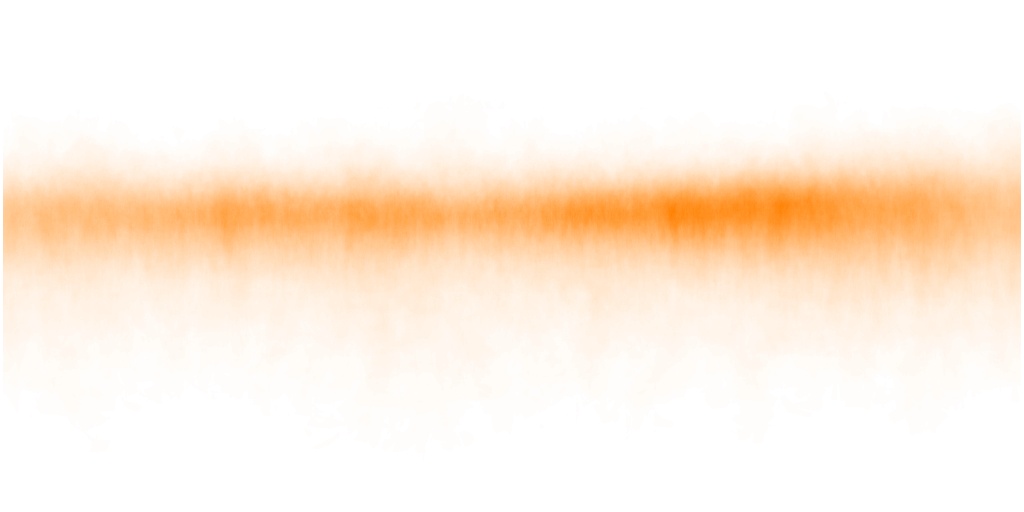}}
        \captionsetup{justification=centering,skip=0pt}
        \caption*{\footnotesize{Person}}
    \end{subfigure}
    \hspace{0.01cm}
    \begin{subfigure}[b]{\imgsize\columnwidth}
        \fbox{\includegraphics[width=\textwidth]{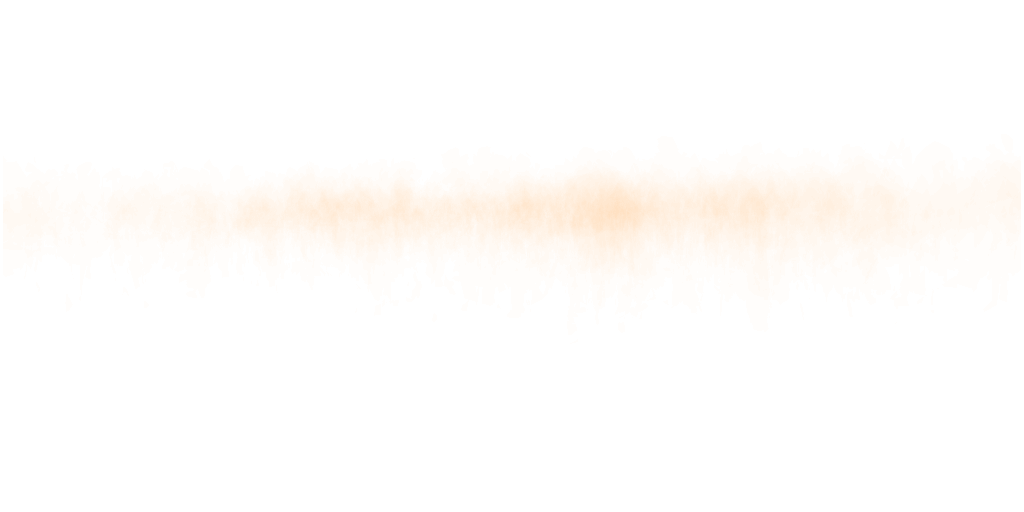}}
        \captionsetup{justification=centering,skip=0pt}
        \caption*{\footnotesize{Rider}}
    \end{subfigure}
    \hspace{0.01cm}
    \begin{subfigure}[b]{\imgsize\columnwidth}
        \fbox{\includegraphics[width=\textwidth]{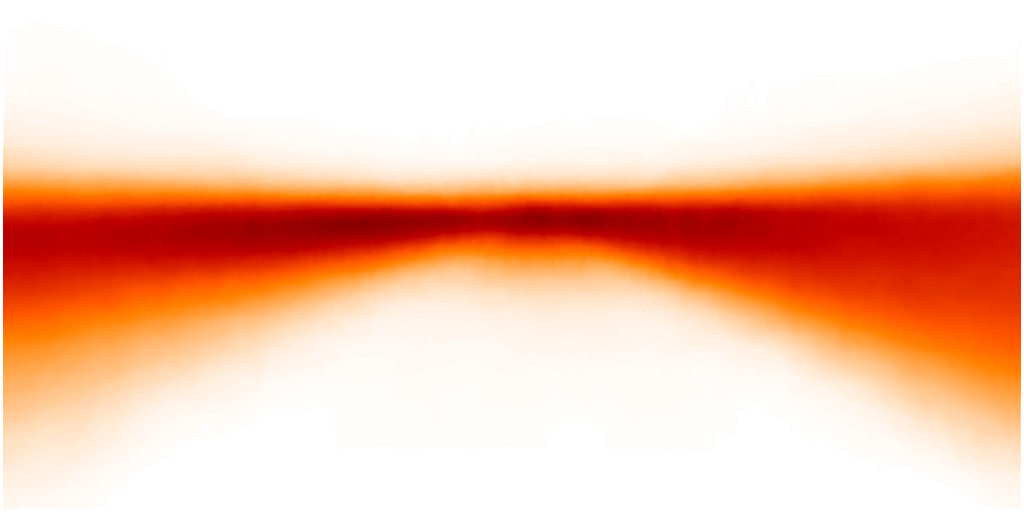}}
        \captionsetup{justification=centering,skip=0pt}
        \caption*{\footnotesize{Car}}
    \end{subfigure}
    \hspace{0.01cm}
    \begin{subfigure}[b]{\imgsize\columnwidth}
        \fbox{\includegraphics[width=\textwidth]{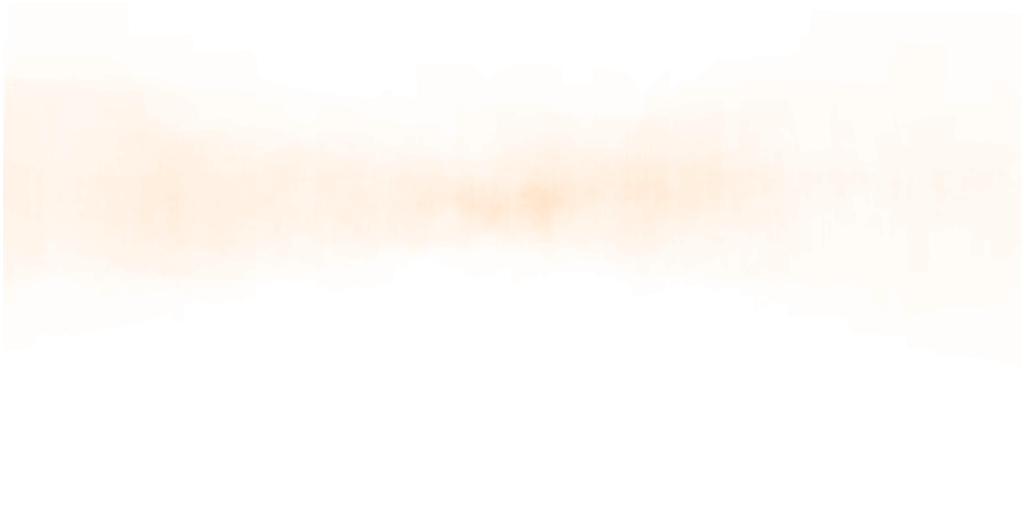}}
        \captionsetup{justification=centering,skip=0pt}
        \caption*{\footnotesize{Truck}}
    \end{subfigure}
    \hspace{0.01cm}
    \begin{subfigure}[b]{\imgsize\columnwidth}
        \fbox{\includegraphics[width=\textwidth]{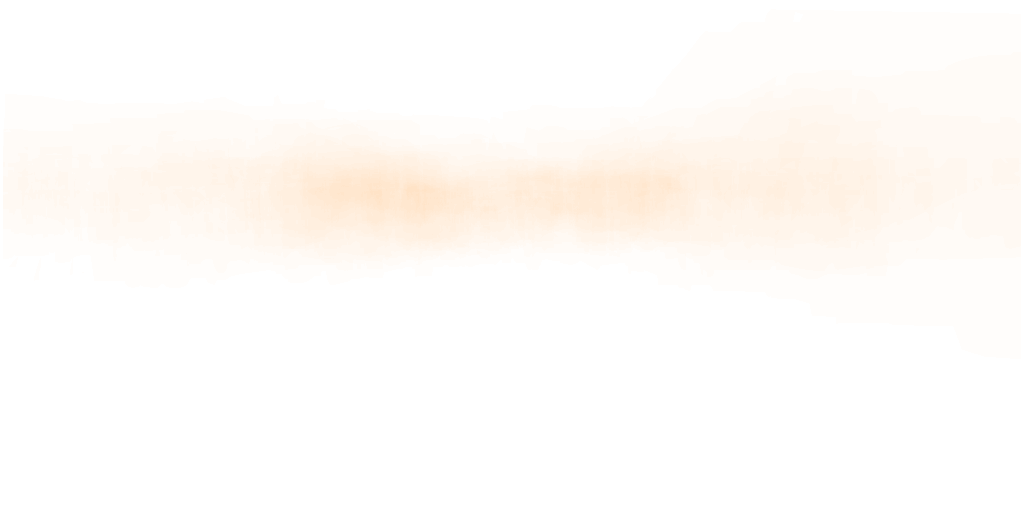}}
        \captionsetup{justification=centering,skip=0pt}
        \caption*{\footnotesize{Bus}}
    \end{subfigure}
    \hspace{0.01cm}
    \begin{subfigure}[b]{\imgsize\columnwidth}
        \fbox{\includegraphics[width=\textwidth]{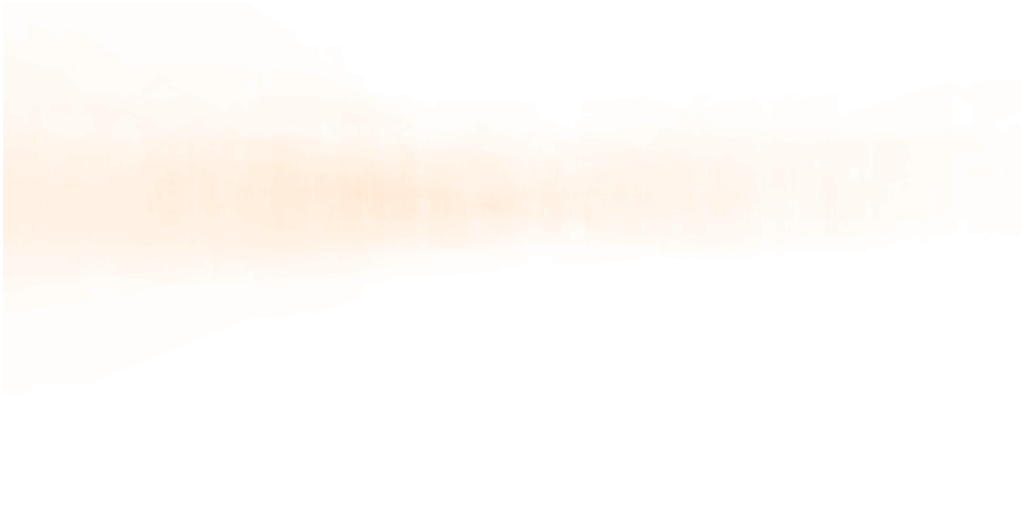}}
        \captionsetup{justification=centering,skip=0pt}
        \caption*{\footnotesize{Train}}
    \end{subfigure}
    \hspace{0.01cm}
    \begin{subfigure}[b]{\imgsize\columnwidth}
        \fbox{\includegraphics[width=\textwidth]{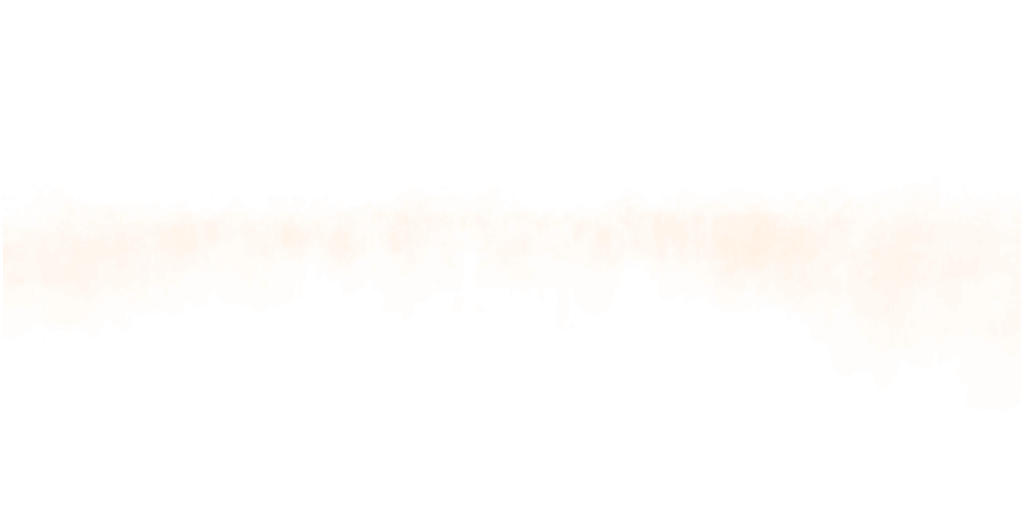}}
        \captionsetup{justification=centering,skip=0pt}
        \caption*{\footnotesize{Motorcycle}}
    \end{subfigure}
    \hspace{0.01cm}
    \begin{subfigure}[b]{\imgsize\columnwidth}
        \fbox{\includegraphics[width=\textwidth]{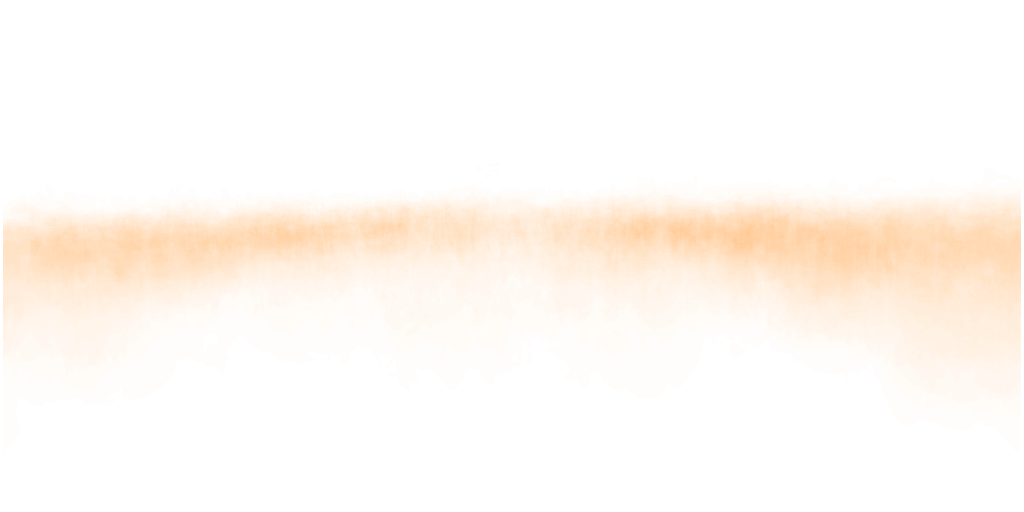}}
        \captionsetup{justification=centering,skip=0pt}
        \caption*{\footnotesize{Bicycle}}
    \end{subfigure}
    %\hspace{12.55cm}
    
    \caption{Spatial distribution of all classes in the Cityscapes training dataset. Dark pixels correspond to more frequent appearance.}
    \label{fig:spatialCS}
\end{figure}

\vspace{0.4cm}
\noindent {\bf Pascal VOC 2012.}
In Table \ref{tab:pascalresults}, we present the results from using our method on the Pascal VOC 2012 dataset. We compare our results to the same four papers as for Cityscapes. We note that our results are competitive, and that we have the strongest performance for two data amounts. There is, however, a significant difference in baselines between the different works, complicating the comparison of the results. In particular, French et al. have a significantly lower baseline, largely because their network is not pre-trained on MSCOCO, resulting in a bigger room for improvement, as shown in the table. We use a network pre-trained on MSCOCO because that is what most existing work is using. Our results can also be compared to those from Ouali et al., who obtain 69.4\% mIoU on Pascal VOC 2012 for 1.5k samples using DeepLabv3 \cite{ouali2020semisupervised}. Our results for 1/8 (or 1323) labelled samples is higher, despite us using fewer labelled samples and a less sophisticated network architecture.
%It is clear that our method is not performing as strongly here as it is for Cityscapes. We obtain the best results for two out of five data amounts, and do not obtain the highest Delta for any data amount.

\begin{table*}[tp!]
    \centering
    \caption{Performance (mIoU) on the Pascal VOC 2012 validation set, results are given from a single run. Results from four previous papers are provided for comparison, all using the same DeepLab-v2 network with a ResNet-101 backbone.}
    \begin{tabular}{ l l l l l l l}
        Labelled samples & 1/100 & 1/50 & 1/20 & 1/8 & 1/4 & Full (10582) \\
        \hline \hline
        
        Baseline & - & 53.2\%\footnotemark & 58.7\%\footnotemark[\value{footnote}] & 66.0\% & 68.3\% & 73.6\% \\
        Adversarial \cite{Hung} & - & 57.2\%\footnotemark[\value{footnote}] & 64.7\%\footnotemark[\value{footnote}] & 69.5\% & 72.1\% & - \\
        Improvement & - & 4.0 & 6.0 & 3.5 & \textbf{3.8} & - \\
        \hline \hline
        
        Baseline & - & 53.2\% & 58.7\% & 66.0\% & - & 73.6\%  \\
        s4GAN \cite{Mittal} & - & 63.3\% & 67.2\% & \textbf{71.4\%} & - & 75.6\% \\
        Improvement & - & 10.1 & 8.5 & 5.4 & - & 2.0 \\
        \hline \hline
        
        Baseline & 33.09\% & 43.15\% & 52.05\% & 60.56\% & - & 72.59\% \\
        French et al. \cite{French}\footnotemark & 53.79\% & 64.81\% & 66.48\% & 67.60\% & - & - \\
         Improvement & \textbf{20.70} & \textbf{21.66} & \textbf{14.48} & \textbf{7.04} & - & - \\
        \hline \hline
        
        Baseline & 45.7\%\footnotemark & 55.4\% & 62.2\% & 66.2\% & 68.7\% & 73.5\% \\
        DST-CBC \cite{Feng} & \textbf{61.6\%}\footnotemark[\value{footnote}]  & 65.5\% & \textbf{69.3\%} & 70.7\% & 71.8\% & - \\
        Improvement & 15.9 & 10.1 & 7.1 & 4.5 & 3.1 & - \\
        \hline \hline
        
        Baseline & 42.47\% & 55.69\% & 61.36\% & 67.14\% & 70.20\% & 74.13\% \\

        Ours & 54.18\% & \textbf{66.15\%} & 67.77\% & 71.00\% & \textbf{72.45\%} & - \\

        Improvement & 11.71 & 10.46 & 6.41 & 3.86 & 2.25 & - \\
        \hline
        
    \end{tabular}
    \label{tab:pascalresults}
\end{table*}{}

Our results are not as strong for the Pascal dataset as they are for Cityscapes. We believe that this is largely because Pascal contains very few classes in each image, usually only a background class and one or two foreground classes. This means that the diversity of masks in ClassMix will be very small, with the same class or classes frequently being selected for mask generation for any given image. This is in contrast to Cityscapes as discussed above. The images in the Pascal dataset are also not that similar to each other. There is no pattern to where in the images certain classes appear, or in what context, unlike Cityscapes. Therefore, pasted objects often end up in unreasonable contexts, which we believe is detrimental for performance. The patterns of where classes appear are made obvious by calculating the spatial distribution of classes, which is visualised for Pascal in Figure \ref{fig:spatialPascal}, and can be compared to Cityscapes in Figure \ref{fig:spatialCS}. In these figures it is clear that the spatial distributions are much less uniform for Cityscapes than for Pascal. We note that in spite of these challenges for the Pascal VOC dataset, ClassMix still performs competitively with previous state of the art.

\newcommand\pascalimgsize{0.085}
\begin{figure}[h!]
    \centering
    
    \setlength{\fboxsep}{0pt}%
    \setlength{\fboxrule}{0.5pt}%
    \begin{subfigure}[b]{\pascalimgsize\textwidth}
        \captionsetup{justification=centering,skip=0pt}
        \fbox{\includegraphics[width=\textwidth]{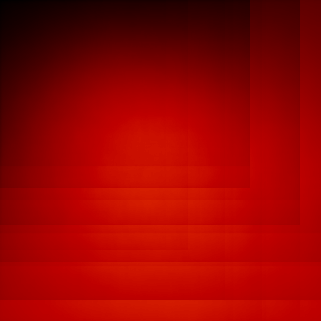}}
        \caption*{\footnotesize{Background}}
    \end{subfigure}
    \hspace{0.01cm}
    \begin{subfigure}[b]{\pascalimgsize\textwidth}
        \fbox{\includegraphics[width=\textwidth]{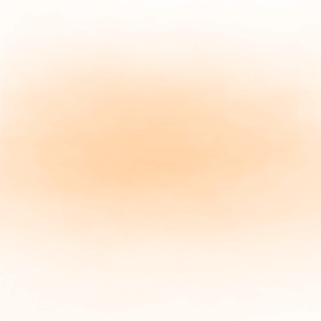}}
        \captionsetup{justification=centering,skip=0pt}
        \caption*{\footnotesize{Aeroplane}}
    \end{subfigure}
    \hspace{0.01cm}
    \begin{subfigure}[b]{\pascalimgsize\textwidth}
        \fbox{\includegraphics[width=\textwidth]{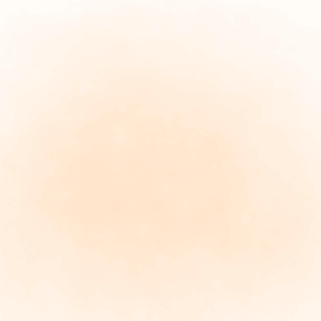}}
        \captionsetup{justification=centering,skip=0pt}
        \caption*{\footnotesize{Bicycle}}
    \end{subfigure}
    \hspace{0.01cm}
    \begin{subfigure}[b]{\pascalimgsize\textwidth}
        \fbox{\includegraphics[width=\textwidth]{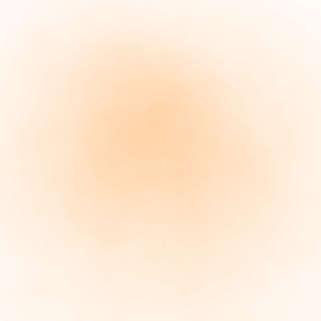}}
        \captionsetup{justification=centering,skip=0pt}
        \caption*{\footnotesize{Bird}}
    \end{subfigure}
    \hspace{0.01cm}
    \begin{subfigure}[b]{\pascalimgsize\textwidth}
        \fbox{\includegraphics[width=\textwidth]{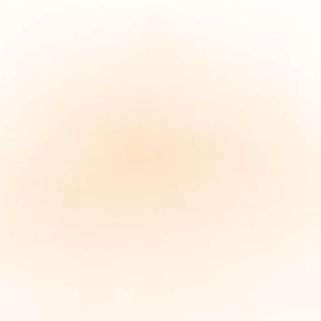}}
        \captionsetup{justification=centering,skip=0pt}
        \caption*{\footnotesize{Boat}}
    \end{subfigure}
    \hspace{0.01cm}
    \begin{subfigure}[b]{\pascalimgsize\textwidth}
        \fbox{\includegraphics[width=\textwidth]{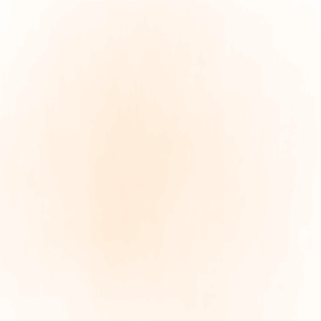}}
        \captionsetup{justification=centering,skip=0pt}
        \caption*{\footnotesize{Bottle}}
    \end{subfigure}
    \hspace{0.01cm}
    \begin{subfigure}[b]{\pascalimgsize\textwidth}
        \fbox{\includegraphics[width=\textwidth]{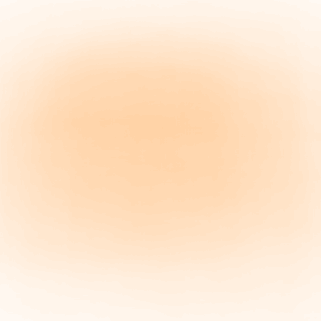}}
        \captionsetup{justification=centering,skip=0pt}
        \caption*{\footnotesize{Bus}}
    \end{subfigure}
    \hspace{0.01cm}
    \begin{subfigure}[b]{\pascalimgsize\textwidth}
        \fbox{\includegraphics[width=\textwidth]{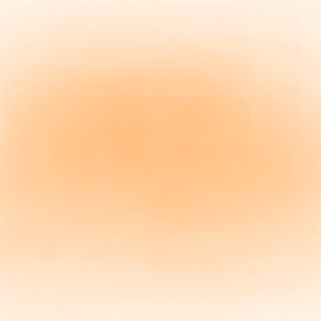}}
        \captionsetup{justification=centering,skip=0pt}
        \caption*{\footnotesize{Car}}
    \end{subfigure}
    \hspace{0.01cm}
    \begin{subfigure}[b]{\pascalimgsize\textwidth}
        \fbox{\includegraphics[width=\textwidth]{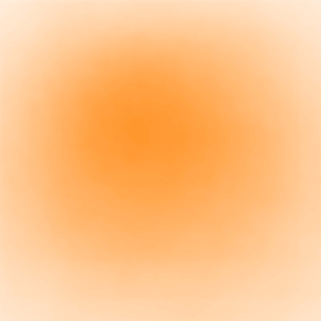}}
        \captionsetup{justification=centering,skip=0pt}
        \caption*{\footnotesize{Cat}}
    \end{subfigure}
    \hspace{0.01cm}
    \begin{subfigure}[b]{\pascalimgsize\textwidth}
        \fbox{\includegraphics[width=\textwidth]{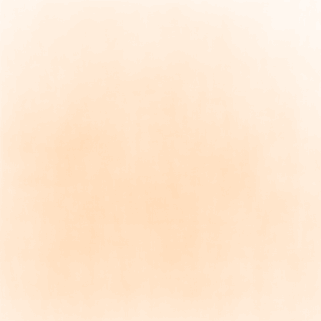}}
        \captionsetup{justification=centering,skip=0pt}
        \caption*{\footnotesize{Chair}}
    \end{subfigure}
    \hspace{0.01cm}
    \begin{subfigure}[b]{\pascalimgsize\textwidth}
        \fbox{\includegraphics[width=\textwidth]{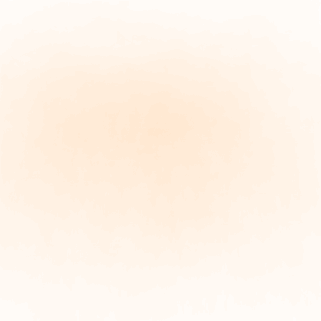}}
        \captionsetup{justification=centering,skip=0pt}
        \caption*{\footnotesize{Cow}}
    \end{subfigure}
    \hspace{0.01cm}
    \begin{subfigure}[b]{\pascalimgsize\textwidth}
        \fbox{\includegraphics[width=\textwidth]{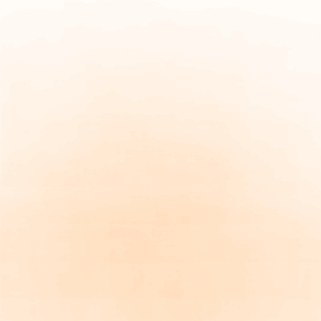}}
        \captionsetup{justification=centering,skip=0pt}
        \caption*{\footnotesize{Dining table}}
    \end{subfigure}
    \hspace{0.01cm}
    \begin{subfigure}[b]{\pascalimgsize\textwidth}
        \fbox{\includegraphics[width=\textwidth]{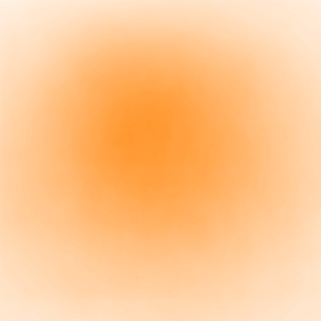}}
        \captionsetup{justification=centering,skip=0pt}
        \caption*{\footnotesize{Dog}}
    \end{subfigure}
    \hspace{0.01cm}
    \begin{subfigure}[b]{\pascalimgsize\textwidth}
        \fbox{\includegraphics[width=\textwidth]{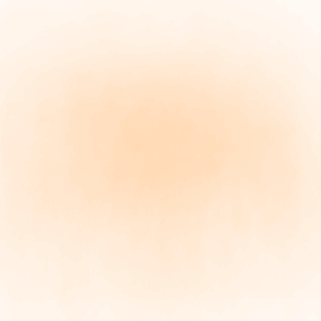}}
        \captionsetup{justification=centering,skip=0pt}
        \caption*{\footnotesize{Horse}}
    \end{subfigure}
    \hspace{0.01cm}
    \begin{subfigure}[b]{\pascalimgsize\textwidth}
        \fbox{\includegraphics[width=\textwidth]{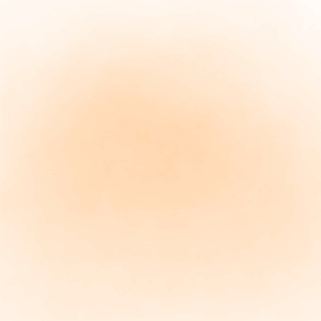}}
        \captionsetup{justification=centering,skip=0pt}
        \caption*{\footnotesize{Motorbike}}
    \end{subfigure}
    \hspace{0.01cm}
    \begin{subfigure}[b]{\pascalimgsize\textwidth}
        \fbox{\includegraphics[width=\textwidth]{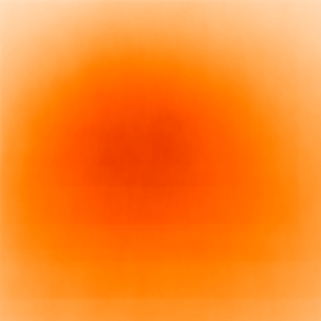}}
        \captionsetup{justification=centering,skip=0pt}
        \caption*{\footnotesize{Person}}
    \end{subfigure}
    \hspace{0.01cm}
    \begin{subfigure}[b]{\pascalimgsize\textwidth}
        \fbox{\includegraphics[width=\textwidth]{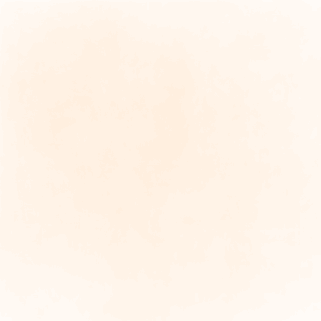}}
        \captionsetup{justification=centering,skip=0pt}
        \caption*{\footnotesize{Potted plant}}
    \end{subfigure}
    \hspace{0.01cm}
    \begin{subfigure}[b]{\pascalimgsize\textwidth}
        \fbox{\includegraphics[width=\textwidth]{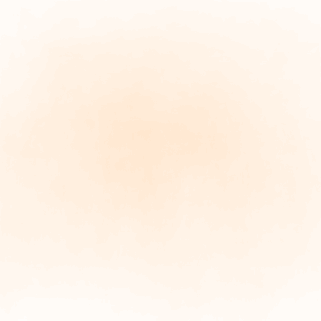}}
        \captionsetup{justification=centering,skip=-0pt}
        \caption*{\footnotesize{Sheep}}
    \end{subfigure}
    \hspace{0.01cm}
    \begin{subfigure}[b]{\pascalimgsize\textwidth}
        \fbox{\includegraphics[width=\textwidth]{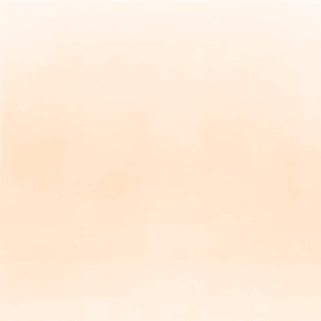}}
        \captionsetup{justification=centering,skip=0pt}
        \caption*{\footnotesize{Sofa}}
    \end{subfigure}
    \hspace{0.01cm}
    \begin{subfigure}[b]{\pascalimgsize\textwidth}
        \fbox{\includegraphics[width=\textwidth]{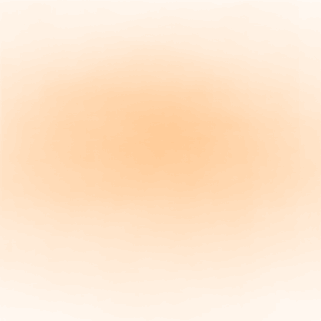}}
        \captionsetup{justification=centering,skip=0pt}
        \caption*{\footnotesize{Train}}
    \end{subfigure}
    \hspace{0.01cm}
    \begin{subfigure}[b]{\pascalimgsize\textwidth}
        \fbox{\includegraphics[width=\textwidth]{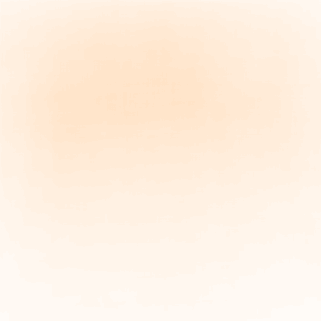}}
        \captionsetup{justification=centering,skip=0pt}
        \caption*{\footnotesize{TV monitor}}
    \end{subfigure}
    %\hspace{12.55cm}
    
    \caption{Spatial distribution of all classes in the Pascal training dataset. Dark pixels correspond to more frequent appearance.}
    \label{fig:spatialPascal}
\end{figure}

\footnotetext[2]{As reported by \cite{Mittal}.}
\footnotetext[3]{Same DeepLab-v2 network but with only ImageNet pre-training and not MSCOCO.}
\footnotetext[4]{Results for 100 (1/106) samples.}

\subsection{Ablation Study}

We investigate our method further by training models with some components changed or removed, in order to see how much those specific components contribute to the performance of the overall algorithm. Additionally, we also experiment with additions %\footnote{Although not ablations in the strict sense, we believe that such changes are still informative, so we include them in this section.} 
to the algorithm, namely adding more augmentations and training for a longer time. Although such additions increase the final performance, they also make comparisons with other existing approaches unfair, which is why these results are not presented in subsection \ref{subsec:results}. The ablation results are presented in Table \ref{tab:Ablation}. All figures are from training a model with 1/8 labelled samples on the Cityscapes dataset, with the same settings as used for the main results except for the part being examined.

\begin{table}[h]
    \centering
    \caption{Ablation study of the proposed method on the Cityscapes dataset. All results are mIoU scores averaged over three runs.}
    \begin{tabular}{ l l } 
        Settings &  mIoU \\
        \hline \hline
        Baseline& 54.84\% \\
        Default SSL & 61.35\% \\ 
        \hline 
        CowMix  & 60.37\% \\
        CutMix & 59.12\% \\
        \hline 
        Pixel-wise threshold  & 58.61\%\\
        Sigmoid ramp up  & 60.58\% \\
        Constant unsupervised weight  & 60.58\% \\
        \hline 
        Squared error loss   & 58.74\% \\
        \hline
        No Pseudo-label & 60.15\% \\
        \hline
        Random crop Baseline & 56.42\% \\
        Random crop & 62.16\% \\
        \hline
        Extra augmentations & 61.85\% \\
        \hline
        80k iterations Baseline & 55.05\% \\
        80k iterations & 62.92\% \\

        \hline
    \end{tabular}
    \label{tab:Ablation}
\end{table}{}

First, we examine the effect of using different mixed sample data augmentations. Apart from ClassMix we try CutMix \cite{CutMix}, as used for semi-supervised semantic segmentation in \cite{French}, and CowMix, introduced by French et al. \cite{MilkingCowMask}. We note that CowMix is very similar to the concurrent FMix, introduced by Harris et al. \cite{harris2020understanding}. As can be seen in Table \ref{tab:Ablation}, ClassMix performs significantly better than both other mixes. That CutMix performs the worst, we attribute to CutMix's masks being less varied than for the other methods. Both ClassMix and CowMix yield flexible masks and we speculate that ClassMix achieves higher results because the masks will follow semantic boundaries to a high degree, giving more natural borders between the two mixed images.

We try three different ways of weighting the unsupervised loss, additional to our default way of weighting it against the proportion of pixels that have a maximum predicted value above a threshold 0.968, as used in \cite{French}. In contrast to this is pixel-wise threshold, where instead all pixels with predicted certainties below the threshold are masked and ignored in the loss. As can be seen in Table \ref{tab:Ablation}, using the pixel-wise threshold significantly lowers the results. We have found that this strategy masks almost all pixels close to class boundaries, as well as some small objects, such that no unsupervised training is ever performed on this kind of pixels, as also noted in \cite{French}.
Sigmoid ramp up increases the unsupervised weight $\lambda$ from 0 to 1 over the course of the first 10k iterations, similarly to what was done in, e.g., \cite{TarvainenMeanTeacher,LaineTemporalEnsembling}. This yields results somewhat lower than in our default solution, and exactly the same results as when keeping the unsupervised weight at a constant 1.

We investigate adjusting the unsupervised loss by changing our default cross-entropy loss with a squared error loss. The loss is summed over the class probability dimension and averaged over batch and spatial dimensions, in keeping with \cite{French}. This yields results considerably lower than when using cross-entropy. 
We also try training with cross-entropy without using pseudo-labels, and instead merely softmax outputs as targets, which also lowers the results. This is likely because entropy minimization, here in the form of pseudo-labelling, helps the network generalize better, as seen in previous works \cite{FixMatch,pseudo-label}. When not using pseudo-labels we also fail to avoid the problem of ``label contamination'', described in Figure \ref{fig:label_contamination}, causing the network to be trained against unreasonable targets near object boundaries.

In our results for Cityscapes, the images are not cropped. Here, however, we investigate randomly cropping both labelled and unlabelled images to $512\times512$. This increases the performance of both baseline and SSL. The reason for this is likely that cropping adds a regularizing effect. The increase from cropping is larger for the baseline than for SSL, which is believed to be because the SSL solution already receives a regularizing effect from ClassMix, leaving less room for improvement.

Adding color jittering (adjusting brightness, contrast, hue and saturation) and Gaussian blurring also improves the results. These extra augmentations are applied as part of the strong augmentation scheme after ClassMix. This introduces more variation in the data, likely increasing the network's ability to generalize. It also makes the strong augmentation policy more difficult, in line with the use of augmentation anchoring.

Training for 80k iterations instead of 40k improves the results significantly, as can be seen in Table \ref{tab:Ablation}. It improves the results more for SSL than for the baseline, which is likely because there is more training data in SSL training, meaning that overfitting is a smaller issue.

\section{Conclusion}

In this paper we have proposed an algorithm for semi-supervised semantic segmentation that uses ClassMix, a novel data augmentation technique. ClassMix generates augmented images and artificial labels by mixing unlabelled samples together, leveraging on the network's semantic predictions in order to better respect object boundaries. We evaluated the performance of the algorithm on two commonly used datasets, and showed that it improves the state of the art. It is also worth noting that since many semantic segmentation algorithms rely heavily on data augmentations, the ClassMix augmentation strategy may become a useful component also in future methods. Finally, additional motivation for the design choices was presented through an extensive ablation study, where different configurations and training regimes were compared.

\vspace{0.4cm}
\noindent {\bf Acknowledgements.} This work was partially supported by the Wallenberg AI, Autonomous Systems and Software Program (WASP) funded by the Knut and Alice Wallenberg Foundation.

{\small
\bibliographystyle{ieee_fullname}
\bibliography{egbib}
}

\end{document}